\documentclass[journal]{IEEEtran}

\usepackage{cite}
\ifCLASSINFOpdf
	\usepackage[pdftex]{graphicx}
	\graphicspath{{../pdf/}{../jpeg/}}
	\DeclareGraphicsExtensions{.pdf,.jpeg,.png}
\else

	\usepackage[dvips]{graphicx}
	\graphicspath{{../eps/}}
	\DeclareGraphicsExtensions{.eps}
\fi

\usepackage[bookmarks=false]{hyperref}
\hypersetup{
    colorlinks,
    linkcolor={black},
    citecolor={black},
    urlcolor={black},
}

\usepackage{amssymb,amsfonts,amsmath}
\usepackage{mathtools}
\usepackage[linesnumbered,ruled,vlined]{algorithm2e}
\usepackage{arydshln}
\usepackage{booktabs}
\usepackage{tikz}
\usetikzlibrary{fit,calc,positioning,decorations.pathreplacing,matrix}
\usetikzlibrary{shapes,arrows}
\usetikzlibrary{calc}
\usepackage{multirow}

\usepackage{xcolor}

\usepackage{enumitem}

\newcommand{\w}{\mathbf{w}}
\newcommand{\smlv}{\mathbf{v}}
\newcommand{\smlh}{\mathbf{h}}
\newcommand{\x}{\mathbf{x}}
\newcommand{\z}{\mathbf{z}}
\newcommand{\y}{\mathbf{y}}
\newcommand{\Bigh}{\mathbf{H}}

\newcommand{\Bigy}{\mathbf{Y}}

\newcommand{\R}{{\mathbb{R}}}
\newcommand{\N}{{\mathbb{N}}}

\newcommand{\cvec}{\mathbf{c}}
\newcommand{\eye}{\mathbf{I}}
\newcommand{\prox}{\mathcal{S}}

\newcommand{\minus}{\scalebox{0.75}[1.0]{$-$}}

\newcommand{\half}{{\textstyle \frac{1}{2}}}

\newcommand{\ip}[2]{\langle #1,#2 \rangle}

\DeclareMathOperator*{\argmax}{arg\,max}
\DeclareMathOperator*{\argmin}{arg\,min}

\begin{document}

\title{Deep Residual Autoencoders for Expectation Maximization-inspired Dictionary Learning}

\author{Bahareh~Tolooshams,~\IEEEmembership{Student Member,~IEEE,}
        Sourav~Dey,~\IEEEmembership{Member}
        and~Demba~Ba,~\IEEEmembership{Member,~IEEE}% <-this % stops a space
\thanks{B. Tolooshams  and D. Ba is with the School of Engineering and Applied Sciences, Harvard University, Cambridge, MA, 02138 USA e-mail: (btolooshams@seas.harvard.edu; demba@seas.harvard.edu)}% <-this % stops a space
\thanks{S. Dey is with Manifold AI, Oakland, CA e-mail:(sdey@manifold.ai)}%
%\thanks{Manuscript received April, 2019; revised March, 2019.}}%
\thanks{Manuscript accepted June, 2020.}}%

\markboth{accepted to the IEEE Transactions on Neural Networks and Learning Systems.}%
{Tolooshams \MakeLowercase{\textit{et al.}}: Constraint Recurrent Sparse autoencoders}


\thispagestyle{empty}

{\bf
\onecolumn
\noindent Preprint Notice:\\

\noindent (c) 2020 IEEE.  Personal use of this material is permitted.  Permission from IEEE must be obtained for all other uses, in any current or future media, including reprinting/republishing this material for advertising or promotional purposes, creating new collective works, for resale or redistribution to servers or lists, or reuse of any copyrighted component of this work in other works.
}

\clearpage
\pagenumbering{arabic} 

\twocolumn
\newpage

\maketitle

\begin{abstract}

We introduce a neural-network architecture, termed the constrained recurrent sparse autoencoder (CRsAE), that solves convolutional dictionary learning problems, thus establishing a link between dictionary learning and neural networks. Specifically, we leverage the interpretation of the alternating-minimization algorithm for dictionary learning as an approximate Expectation-Maximization algorithm to develop autoencoders that enable the simultaneous training of the dictionary and regularization parameter (ReLU bias). The forward pass of the encoder approximates the sufficient statistics of the E-step as the solution to a sparse coding problem, using an iterative proximal gradient algorithm called FISTA. The encoder can be interpreted either as a recurrent neural network or as a deep residual network, with two-sided ReLU non-linearities in both cases. The M-step is implemented via a two-stage back-propagation. The first stage relies on a linear decoder applied to the encoder and a norm-squared loss. It parallels the dictionary update step in dictionary learning. The second stage updates the regularization parameter by applying a loss function to the encoder that includes a prior on the parameter motivated by Bayesian statistics. We demonstrate in an image-denoising task that CRsAE learns Gabor-like filters, and that the EM-inspired approach for learning biases is superior to the conventional approach. In an application to recordings of electrical activity from the brain, we demonstrate that CRsAE learns realistic spike templates and speeds up the process of identifying spike times by $\textbf{900}$x compared to algorithms based on convex optimization.
\end{abstract}


% \fi
\IEEEpeerreviewmaketitle

%%%%%%%%%%%%%%%%%%%%%%%%%%%%%%%%%%%%%%%%%%%%%%%%%%%%%%%%%%%%%%%%%%%%%%
\section{Introduction}

\IEEEPARstart{S} PARSE coding has become a popular method for extracting features from data. Sparse coding problems assume the data $\y$ can be represented as a sparse linear combination of columns (features) of a matrix $\Bigh$, termed a dictionary. Given the dictionary $\Bigh$, methods such as orthogonal matching pursuit~\cite{Chen1989OrthogonalLS} and basis pursuit~\cite{Chen1998AtomicDB} find the sparse representation. The Lasso is an alternative method to estimate the sparse code by penalizing the $\ell_1$-norm of the linear regression coefficients~\cite{TibshiraniRobert1996RSaS}. The performance of this regression highly depends on the choice of the regularization parameter. The sparse coding problem also has connections to Bayesian statistics, where Monte-Carlo Expectation-Maximization (EM)~\cite{DempsterAP1977MLfI} can be used for estimation of hyperparameters such as the regularization parameter~\cite{ZhuXinFeng2011LSLa,ParkTrevor2008TBL}. In this line of thought, treating the sparse codes as continuous mixtures of Gaussians makes it possible to obtain Gibbs samples from the missing data of interest, namely the sparse codes and the hidden variance of each of the Gaussians in the scale mixture. The M-step uses the Gibbs samples to approximate the intractable expectations necessary to update the hyperparameters~\cite{ParkTrevor2008TBL}. As with most Monte-Carlo methods, the Gibbs sampling step can be computationally expensive, particularly when the sparse codes are high-dimensional.

%\IEEEpubidadjcol

The dictionary in the sparse coding problem can be analytical (e.g., discrete wavelet transform~\cite{Mallat1989ATF}), have random structures (e.g., Gaussian random matrix), or be learned from data for a better adaptation. In the signal processing literature, dictionary learning is the de-facto method for learning the sparse representations in an unsupervised manner. The method of optimal directions~\cite{EnganK1999Mood} and K-SVD~\cite{AharonM2006rKAA} are two examples of popular dictionary learning algorithms. In convolutional dictionary learning (CDL), which has drawn attention in signal and image processing due to its ability to produce shift-invariant sparse representations, $\Bigh$ has a block-Toeplitz structure consisting of the concatenation of a finite number of Toeplitz matrices, each corresponding to a convolutional filter (atom)~\cite{garcia-2018-convolutional,PapyanV2017CNNA}. CDL is a non-convex optimization problem. A popular method to circumvent this non-convexity is to alternate between a convolutional sparse coding (CSC) step and a convolutional dictionary update step until a convergence criterion is met, a procedure referred to as \emph{alternating minimization}~\cite{Agarwal2016LearningSU}. The state-of-the-art algorithms solve the CSC step through variants of ADMM~\cite{Boyd2011DistributedOA}. While efficient, these algorithms lack a scalable infrastructure to solve the CSC step in parallel for thousands of examples.

A recent line of work has sought to draw connections between autoencoders (AE)s and dictionary learning~\cite{SreterHillel2018LCSC,Rolfe2013DiscriminativeRS,Makhzani2013kSparseA}. Expressing sparse coding and dictionary learning as a feed-forward neural network lets us take advantage of the parallelism offered by GPUs to speed up learning. In this line of work, the \emph{encoder} defines a mapping from the input data space to a sparse vector similar to the sparse coding step, and the \emph{decoder}, imitating the dictionary update step, uses a set of filters (dictionary) to reconstruct the input data from the approximated sparse codes. We showed in our previous work~\cite{TolooshamsBahareh2018SCDL} that, unless the weights of the encoder and decoder are tied, the architectures from~\cite{SreterHillel2018LCSC} and~\cite{Rolfe2013DiscriminativeRS} do not, strictly speaking, perform dictionary learning. That is, for the weights of an AE to be interpretable as convolutional filters in a CDL problem, encoder and decoder weights must necessarily be tied. Besides, we showed using simulated data that the AE introduced in~\cite{SreterHillel2018LCSC} is not able to learn the ground-truth underlying convolutional dictionary because of a) encoder and decoder weights are not tied, and  b) the codes produced by the encoder are not sparse enough to enable dictionary learning. In the context of dense (as opposed to convolutional) dictionary learning, the $k$-sparse AE~\cite{Makhzani2013kSparseA} imposes the afore-mentioned constraint and uses a variant of iterative hard-thresholding~\cite{Blumensath2008IterativeHT} in its encoder to ensure that the encoder outputs sparse codes. The efficacy of limiting the code to be $k$-sparse through hard-thresholding is not apparent in the convolutional case. Additionally, in all of the afore-mentioned works, a principled method for training the regularization parameters (ReLU biases) seems to be missing.

The next section, Section~\ref{sec:contri}, summarizes our contributions. Section~\ref{sec:GEN} introduces notation and the generative model. We review classical CDL in Section~\ref{sec:DL}. Section~\ref{sec:EM} discusses the EM approach to dictionary learning. We introduce a deep residual AE architecture motivated by EM in Section~\ref{sec:crsae}. In Section~\ref{sec:EXP}, we compare the EM-inspired deep residual AE to existing algorithms for CDL and apply it to spike sorting and image denoising. We conclude in Section~\ref{sec:CONC}.

\vspace*{-1mm}
\section{Summary of Contributions}\label{sec:contri}

\noindent The contributions of this paper are

\vspace*{1mm}
\noindent {\bf \underline{A deep residual AE for CDL}}: We propose an AE architecture for CDL (Figure~\ref{fig:blockdiagram}) that is an extension of that introduced in~\cite{TolooshamsBahareh2018SCDL}. The encoder in this architecture is a variant of residual~\cite{HeKaiming2015DRLf} and unrolled recurrent networks~\cite{gregor2010learning,Rolfe2013DiscriminativeRS} (Figure~\ref{fig:resnet}).

\vspace*{1mm}
\noindent {\bf  \underline{A training algorithm \emph{inspired} by EM}}: We extend our work from~\cite{TolooshamsBahareh2018SCDL} and provide a prescription for simultaneously training the filters and biases from the generative model that underlies CDL. This procedure is motivated by EM and Bayesian statistics. The encoder, mimics the sparse coding step in dictionary learning, and approximates the E-step of EM. The M-step is a two-stage back-propagation procedure (Figure~\ref{fig:backprop}). In the first stage, we perform back-propagation through the AE, a step that parallels the dictionary update step in dictionary learning. In the second stage, we perform back-propagation through the encoder using a loss function motivated by Bayesian statistics. This last step, which allows us to train the regularization parameter (ReLU bias) (Figure~\ref{fig:lambda}), does not have an equivalent in dictionary learning. We show in Section~\ref{sec:image} that training the bias with this EM-inspired approach results in a network that performs image denoising better than one that trains it using the conventional approach in deep learning, namely by minimizing a global reconstruction loss (Table~\ref{tab:psnr}).

\vspace*{1mm}
\noindent {\bf \underline{An architecture that is interpretable and robust}}: We demonstrate that, when trained on data simulated from a generative model with known filters, this AE architecture motivated by CDL can a) successfully learn ground-truth filters (Figures~\ref{fig:H_sim_err} and~\ref{fig:H_sim_learned}), and b) is robust to noise at a range of signal-to-noise ratio (SNR) (Figure~\ref{fig:SNR}). That is, ours is an AE architecture for which the filters are interpretable as coming from a generative model. This is corroborated by experiments on natural images, absent in~\cite{TolooshamsBahareh2018SCDL}, which show that the architecture learns Gabor-like filters (Figure~\ref{fig:filters}).

\vspace*{1mm}
\noindent {\bf \underline{An application to spike sorting}}: We show that the architecture can perform source separation when the support of the sources is much smaller than the observed data. We demonstrate this ability by separating the activity of neurons in recordings of electrical activity from the brain of rats (Figure~\ref{fig:H_real}). The encoder from the architecture performs spike sorting, namely,  identifies the location of action potentials from individual neurons in the recordings (Figure~\ref{fig:miss_false}).

\vspace*{1mm}
\noindent {\bf \underline{A demonstration of significant computational gains}}: We extend our work from~\cite{TolooshamsBahareh2018SCDL} and demonstrate that this approach, which can readily employ off-the-shelf tools for training networks on GPUs, is $5$x faster than a state-of-the-art CDL algorithm based on convex optimization (Figure~\ref{fig:speed_analysis}), and performs spike sorting on hours of raw electrophysiological recordings $900$x faster than existing optimization-based methods (Table~\ref{tab:speed}, Spike Sorting).

\vspace*{1mm}
\noindent {\bf \underline{A neural network for image denoising}}: We demonstrate that our framework can successfully denoise images, and rivals state-of-the-art methods with fewer parameters. In addition, we explore the effects of encoder depth, weight sharing between encoder and decoder, as well as different methods of training biases, on denoising performance (Table~\ref{tab:psnr}). This application was not demonstrated in~\cite{TolooshamsBahareh2018SCDL}.
%%%%%%%%%%%%%%%%%%%%%%%%%%%%%%%%%%%%%%%%%%%%%%%%%%%%%%%%%%%%%%%%%%%%%%
%\vspace*{-8mm}
\section{Generative Model}
\label{sec:GEN}
\vspace{-1mm}

\subsection{Notation}
\vspace{-1mm}
\noindent We follow the notational conventions summarized in Table~\ref{tab:notations}.

\begin{table}[!ht]  
%\vspace{-4mm}
\caption{Notation and conventions}
\label{tab:notations}
\vspace*{-2mm}
  \centering
  \begin{tabular}{cl}
    Symbol  & Description \\ \midrule
    $\Bigh$ & Matrix (upper-case bold letters)\\ \midrule
    $\smlh$ & Vector (lower-case bold letters)\\ \midrule
    $h_{c}[i]$ & $i^{\text{th}}$ element of the vector $\smlh_c$  \\ \midrule
    $\z_t$ & Vector $\z$ at the $t^{\text{th}}$ iteration of an iterative algorithm  \\ \midrule
    $z_{t}[n]$ & $n^{\text{th}}$ element of $\z_t$  \\ \midrule
    $\Bigh^{\text{T}}$ & Transpose of $\Bigh$  \\ \midrule
    $\smlh^{\text{T}}$ &  Transpose of $\smlh$  \\ \midrule
    $\y^j$ & $j^{\text{th}}$ training example (window of data)  \\ \midrule
     $\|\x\|_p$ & The $\ell_p$-norm of vector $\x$   \\ \midrule
    $\sigma_\text{max}(\Bigh)$ & Maximum eigenvalue of the matrix $\Bigh$  \\ \midrule
    $\eye$ & The identity matrix \\ \midrule
    $\mathbb{I}$ & The indicator function \\ \midrule 
    $*$ & Linear convolution \\ \midrule 
    $i.i.d.$ & Independent and identically distributed \\ \midrule 
    $\mathcal{N}(m,\sigma^2)$ & Gaussian distribution with mean $m$ and variance $\sigma^2$ \\  \bottomrule
   \\
  \end{tabular}
\vspace{-4mm}
\end{table}

We focus on signals with a one-dimensional domain, namely time. A generalization to signals such as images or videos is simply a matter of replacing one-dimensional convolutions with multi-dimensional ones.

\vspace*{-4mm}
\subsection{Continuous-time Model}
\noindent Consider a set of continuous-time filters $(h_c(t))_{c=1}^C$ localized in time where $C$$\in$$\N^+$ is the number of filters. Let $y(t)$ be the continuous-time signal that is the linear mixture of time-shifted and scaled versions of $(h_c(t))_{c=1}^C$, and formally defined as
\vspace*{-2mm}
\begin{equation}\label{eq:CTconv}
y(t) = \sum_{c=1}^C \sum_{i=1}^{N_c} x_{c,i}h_c(t-\tau_{c,i}) + v(t) \quad\textrm{for }t\in [0,T_0),
\end{equation}
\vspace*{-3mm}

\noindent where $v(t) \stackrel{i.i.d.}{\sim} \mathcal{N}(0,\sigma^2)$ is additive noise, $N_c \in \N$ is the number of appearances of filter $c$ in the signal, and $x_{c,i} \in \R$ and $\tau_{c,i} \in \R^+$ encode, respectively, the amplitude and position of the $i^{\text{th}}$ appearance of filter $c$ in the signal.
%%%%%%%%%%%%%%%%%%%%%%%%%%%%%%%%%%
%\vspace*{-2.2mm}
\subsection{Discrete-time Model}
\noindent  Assuming $y(t)$ is sampled at the rate $f_s$, its discrete-time version $y[n]$ is given by
\begin{equation}
\label{eq:DTconv}
\begin{aligned}
y[n] = \sum_{c=1}^C \sum_{i=1}^{N_c} x_{c,i}h_c[n-&n_{c,i}] + v_n = \sum_{c=1}^C h_c[n] \ast x_c[n] + v[n]\\
&\text{for } n = 1,\ldots,\left\lfloor \frac{T_0}{f_s}\right \rfloor, n_{c,i} = \left \lfloor \frac{\tau_{c,i}}{f_s}\right \rfloor,
\end{aligned}
\end{equation}
\noindent where $x_c[n] = \sum_{i=1}^{N_c} x_{c,i} \delta [n-n_{c,i}]$ and $\{h_c[n]\}_{n=0}^{K-1}$ is the discrete-time analog of $h_c(t)$, for $c=1,\ldots,C$. We can express Eq.~(\ref{eq:DTconv}) in linear-algebraic form as follows
\vspace*{-1.4mm}
\begin{equation}
\label{eq:matconv}
\y = \begin{bmatrix} \Bigh_1 | \cdots | \Bigh_C \end{bmatrix} \begin{bmatrix} \x_1 \\ \vdots \\ \x_C \end{bmatrix} + \smlv = \Bigh \x + \smlv,
\end{equation}
\noindent where $\y$$\in \R^M$, and for $c=1,\ldots,C$, $\x_c = [x_c[0],\ldots,x_c[M-K]]^{\text{T}} \in \R^{N_e=M-K+1}$, $\smlh_c = [h_c[0],\ldots,h_c[K-1]]^{\text{T}} \in \R^K$, $\Bigh_c \in \R^{M \times N_e}$ is the matrix whose columns consist of delayed versions of the filter $\smlh_c$. The matrix $\Bigh \in \R^{M \times N_eC}$ is the concatenation of convolutional operators that, in practice, we do not need to store explicitly, and $\x \in \R^{N_eC}$.

The model in Eq.~(\ref{eq:matconv}) has promising applications in signal and image processing, where the signal of interest is the linear superposition of translated discrete-time filters or templates. Examples are dictionary learning~\cite{garcia-2018-convolutional} and image representation by learned features~\cite{Zeiler2010DeconvolutionalN}. Similar to~\cite{Zeiler2010DeconvolutionalN}, we assume that the support of the filters are much smaller than the size of the signal ($K$$\ll$$M$$=$$\left\lfloor \frac{T_0}{f_s}\right \rfloor$ (Assumption I)). In addition, in dictionary learning~\cite{garcia-2018-convolutional} or sparse coding problems~\cite{Donoho2197}, $\x$ is assumed to be a sparse vector (Assumption II), $\sum_{c=1}^C N_c \ll N_eC$, resulting in $\x_c$ being an $N_c$-sparse vector. Both of these are plausible assumptions in spike sorting~\cite{lewicki1998review}, where $y[n]$ is a recording of extracellular voltage from the brain, the filters represent action potentials from an ensemble of neurons, and the firing rate of the neurons is assumed to be much smaller than the duration of the recording $M$.

In the case of recordings from an array of sensors, we assume that each of the sensors follows the model of Eq.~(\ref{eq:CTconv}). In practice, due to memory constraints, we divide $\y$ into $J$ windows, each of size $N$. As we detail in Section~\ref{sec:EXP}, this also lets us take advantage of batch-based methods for training neural networks. Following Assumptions I \& II, we let $K \ll N \ll M$. For each example $j$, Eq.~(\ref{eq:matconv}) becomes
\vspace*{-2mm}
\begin{equation}\label{eq:conv}
\y^j = \Bigh \x^j + \smlv^j \quad \textrm{for } j = 1,\ldots,J,
\end{equation}
where $\Bigh \in \R^{N \times N_eC}$, $N_e$$=$$N$$-$$K$$+$$1$, and $\x^j$ is $N_c$-sparse. 
\vspace*{-4mm}
\section{Classical Dictionary Learning}\label{sec:DL}
\noindent Sparse coding problems assume that the data $\y^j \in \R^N$ can be represented as a linear combination of a few columns of the matrix $\Bigh$, termed a dictionary. Given a set of signals $\{\y^j\}_{j=1}^J$ and the filters $\Bigh$, sparse coding attempts to find the sparsest representation $\x^j$ by solving the following problem~\cite{Donoho2197}
\vspace*{-2mm}
\begin{equation}\label{eq:P0}
\begin{aligned}
(P_0): \quad \min_{\x^j} \|\x^j\|_0 \text{ s.t. }\Bigh \x^j = \y^j.
\end{aligned}
\vspace*{-1.5mm}
\end{equation}
\noindent  $(P_0)$ is a  non-convex optimization problem. Basis pursuit~\cite{Chen1998AtomicDB} is a convex relaxation of $(P_0)$ that instead solves
\begin{equation}\label{eq:P1}
\begin{aligned}
(P_1): \quad \min_{\x^j} \|\x^j\|_1 \text{ s.t. }\Bigh \x^j = \y^j.
\end{aligned}
\vspace*{-1.0mm}
\end{equation}
\vspace*{-4.0mm}

\noindent  In the case of a signal $\y^j = \Bigh \x^j + \smlv^j$ observed in the presence of bounded noise $\smlv^j$ such that $\|\smlv^j\|_2 \leq \epsilon$, $(P_1)$ can be extended as follows~\cite{Cands2005StableSR}
\vspace*{-1mm}
\begin{equation}\label{eq:P2}
\begin{aligned}
(P_2): \quad \min_{\x^j} \|\x^j\|_1 \text{ s.t. }\|\y^j - \Bigh \x^j\|_2 \leq \epsilon.
\end{aligned}
\end{equation}

\noindent Given the dictionary, $(P_2)$ finds the sparse codes given noisy observations. In source separation problems such as spike sorting, the filters are unknown, and so must be learned along with the codes. CDL assumes that the dictionary $\Bigh \in \R^{N \times N_eC}$ has block-Toeplitz structure and is a linear operator that performs convolution with the filters $\{\smlh_c\}_{c=1}^C$. The goal is to find these filters by solving an extension of $(P_2)$ given by
\vspace*{-2mm}
\begin{equation}\label{eq:CDL}
\begin{aligned}
\quad \min_{\{\x^j\}_{j=1}^J, \{\smlh_c\}_{c=1}^C} \sum_{j=1}^J \|\x^j\|_1 \text{ s.t. }&\|\y^j - \Bigh \x^j\|_2 \leq \epsilon\\
&\|\smlh_c\|_2 = 1 \quad \textrm{for } c = 1,\ldots, C,
\end{aligned}
\end{equation}
\noindent where the constraint on the filters is to avoid scaling ambiguities. Solving for the sparse codes $\{\x^j\}_{j=1}^J$ and the filters $\{\smlh_c\}_{c=1}^C$ simultaneously in Eq.~(\ref{eq:CDL}) is a non-convex optimization problem. The alternating-minimization method solves Eq.~(\ref{eq:CDL}) in two stages: given an initial estimate of the filters, the algorithm alternates between a CSC step to estimate sparse codes $\{\x^j\}_{j=1}^J$ and a dictionary update step to estimate the filters $\{\smlh_c\}_{c=1}^C$ given the newly-estimated sparse codes~\cite{Agarwal2016LearningSU}. In the case of dense (as opposed to convolutional) dictionary,~\cite{Agarwal2016LearningSU} shows that the alternating minimization algorithm converges to the true dictionary whenever the dictionary satisfies RIP~\cite{Cands2008TheRI}.

%%%%%%%%%%%%%%%%%%%%%%%%%%%%%%%%%%
\vspace*{-4mm}
\subsection{Convolutional Sparse Coding Update}\label{sec:CSCU}

\noindent Given the filters $\Bigh$, the CSC step is separable over the $J$ examples. We can solve for the $j^{\text{th}}$ sparse code $\x^j$ using the unconstrained form of $(P_2)$ given by
\vspace*{-2mm}
\begin{equation}\label{eq:csc_update}
\begin{aligned}
\min_{\x^j} \frac{1}{2}\|\y^j - \Bigh \x^j\|_2^2 +\lambda \|\x^j\|_1,
\end{aligned}
\end{equation}

\noindent where $\lambda > 0$ is a regularization parameter that depends on $\epsilon$ and encourages sparsity. State-of-the-art CSC algorithms use ADMM to solve Eq.~(\ref{eq:csc_update})~\cite{wohlberg-2014-efficient}. At present, convex optimization algorithms, such as ADMM, cannot be deployed easily on GPUs to enable the computation of the solution to Eq.~(\ref{eq:csc_update}) for all $J$ examples in parallel. In section~\ref{sec:crsae}, we introduce an architecture that allows us to solve the CSC problem for all $J$ examples in parallel using GPUs.

%%%%%%%%%%%%%%%%%%%%%%%%%%%%%%%%%%
\vspace*{-2.5mm}
\subsection{Convolutional Dictionary Update}\label{sec:CDU}

\noindent Given the sparse codes, the filters are updated as follows
\vspace*{-1mm}
\begin{equation}\label{eq:dictionary_update}
\min_{\{\smlh_c\}_{c=1}^C} \sum_{j=1}^J \frac{1}{2} \|\y^j-\Bigh  \x^j\|_2^2 \text{ s.t. } \|\smlh_c\|_2 = 1 \quad \textrm{for } c = 1,\ldots, C.
\end{equation}
\noindent 
Unlike the CSC step, the dictionary update is \emph{not} parallelizable over the $J$ examples, which makes it computationally expensive. The work in~\cite{garcia-2018-convolutional} has proposed various methods to solve Eq.~(\ref{eq:dictionary_update}), of which we briefly describe two. The first solves Eq.~(\ref{eq:dictionary_update}) using ADMM by introducing a consensus variable \cite{garcia-2018-convolutional, Boyd2011DistributedOA}. The second takes advantage of the symmetry of the convolution and solves a problem equivalent to Eq.~(\ref{eq:dictionary_update}) using gradient-based methods in the DFT-domain~\cite{garcia-2018-convolutional}. In both of these methods for updating the filters, the regularization parameter is treated as a hyperparameter to be tuned through cross-validation. In the next section, we propose an approach inspired by Bayesian statistics and EM to estimate both the filters $\Bigh$ and regularization parameter $\lambda$.

%%%%%%%%%%%%%%%%%%%%%%%%%%%%%%%%%%%%%%%%%%%%%%%%%%%%%%%%%%%%%%%%%%%%%%
\vspace*{-2mm}
\section{Expectation-Maximization-inspired Dictionary Learning and Parameter Estimation}
\label{sec:EM}

\noindent The objective is to estimate the sparse codes, filters, and regularization parameter. We explain how this is achieved using an approximation to the EM algorithm in a Bayesian generative setting. The approximate EM algorithm motivates the AE architecture proposed in section~\ref{sec:crsae} that is able to learn both the filters and regularization parameter in classical CDL.

\vspace{-4mm}
\subsection{Bayesian Generative Model}

\noindent In the remainder of the treatment, we drop the superscript $j$ from Eq.~(\ref{eq:conv}) to simplify the notation. This yields
\vspace{-2mm}
\begin{equation}\label{eq:generative_model}
\y = \Bigh \x+ \smlv,
\end{equation}
\noindent where we assume $\smlv \sim \mathcal{N}(\textbf{0},\sigma^2\eye)$. Hence, given the sparse codes, the filters, and $\sigma^2$, the data are distributed according to a multivariate Gaussian distribution. That is
\begin{equation}\label{eq:likelihood}
\y\mid \x,\Bigh ; \sigma^2 \sim \mathcal{N}(\Bigh \x,\sigma^2\eye),
\end{equation}
\noindent resulting in the data likelihood
\vspace{-1mm}
\begin{equation}\label{eq:likelihood}
P_{\Bigy}(\y\mid \x,\Bigh;\sigma^2) = \frac{1}{(2\pi\sigma^2)^{\frac{N}{2}}}\; e^{\frac{-1}{2\sigma^2}\|\y - \Bigh \x\|_2^2}
\end{equation}
\vspace{-3mm}

\noindent To encourage sparsity, we assume that conditioned on $\lambda$, each sparse code $\x_c$, $c=1,\ldots,C$, is drawn according to distribution whose density is the product of one-dimensional (1D) $i.i.d.$ Laplace probability density functions. That is
\vspace{-1mm}
\begin{equation}\label{eq:prior}
	P(\x_c \mid \lambda) = \prod_{k=1}^{N_e} \frac{\lambda}{2} e^{-\lambda |x_{c}[k]|} = \left(\frac{\lambda}{2}\right)^{N_e} e^{-\lambda \|\x_c\|_1}
\end{equation}
\vspace{-2mm}

\noindent We assume the marginal prior on the filters $P(\Bigh)$ is non-informative, and that $\lambda$ follows a Gamma prior
\begin{equation}\label{eq:gammaprior}
\begin{aligned}
P(\lambda) = \frac{\delta^r}{\Gamma{(r)}} \lambda^{r-1} e^{-\delta\lambda},
\end{aligned}
\end{equation}
\noindent where $r$, and $\delta$ are the shape and rate parameters of the density, respectively. Hence, $E[\lambda] = \frac{r}{\delta}$.

We are interested in the so-called complete-data likelihood, i.e., the joint distribution $P(\y, \x,\Bigh,\lambda ; \sigma^2)$ under the prior
\begin{equation}
	P(\x, \Bigh, \lambda) = \prod_{c=1}^{C} \left(\frac{\lambda}{2}\right)^{N_e} e^{-\lambda \|\x_c\|_1} P(\Bigh) P(\lambda),
\end{equation}
assuming $\sigma^2$ is known, and a non-informative (flat) prior $P(\Bigh)$ on $\Bigh$.

\noindent The log of the complete-data likelihood is of the form
\begin{equation}\label{eq:logpos}
\begin{aligned}
&\log{P(\y, \x, \Bigh, \lambda; \sigma^2)}\\
&=\log{P(\y \mid \x, \Bigh, \lambda; \sigma^2)} + \log{P(\x,\Bigh,\lambda)} + \text{const.}\\
&= -\frac{1}{2\sigma^2}\|\y - \Bigh \x\|^2_2  - \lambda \|\x\|_1\\
&+ C\log{P(\lambda)} + N_eC \log{\lambda} + \text{const.},
\end{aligned}
\end{equation}
\noindent where const. contains the terms that do not depend on $\x$, $\Bigh$ or $\lambda$, including the flat prior on $\Bigh$.

\vspace{-5mm}
\subsection{EM Algorithm}

\noindent The EM algorithm is an algorithm to maximize the marginal log-likelihood $\log{P(\y,\Bigh, \lambda ; \sigma^2)}$ with respect to (w.r.t.) $\Bigh$ and  $\lambda$. Because the marginal likelihood is typically not available in closed form, EM introduces the hidden variable or missing data $\x$ and operates instead on the complete-data log-likelihood $\log{P(\y,\x,\Bigh, \lambda;\sigma^2)}$. The algorithm iteratively alternates between an E-step and M-step, until convergence. 

\vspace*{1mm}
\noindent \underline{\textbf{E-step}}: Let $\theta = (\Bigh,\lambda)$. At iteration $l$, the E-step computes the function $\mathcal{Q}(\theta|\theta^{(l-1)})$, which is the expectation of the complete-data likelihood w.r.t. the posterior of $\x$ given the data $\y$ and the parameters $\Bigh^{(l\minus1)}$ and $\lambda^{(l\minus1)}$
\begin{equation}\label{eq:qfunc}
\mathcal{Q}(\theta|\theta^{(l-1)}) = \mathbb{E}_{\x}[\log{P(\y \mid \x, \Bigh, \lambda; \sigma^2)} | \y, \Bigh^{(l-1)},\lambda^{(l-1)};\sigma^2].
\end{equation}

Let $\x^{(l)}$ be the maximum a posteriori (MAP) estimate of $\x$ at iteration $l$, i.e., the posterior mode, given by
\vspace{-2mm}
\begin{equation}\label{eq:Estep1}
\begin{aligned}
\x^{(l)} &= \argmax_{\x} \log{P(\x \mid \y ,\Bigh^{(l\minus1)},\lambda^{(l\minus1)};\sigma^2)}\\
&= \argmin_{\x} \ \frac{1}{2\sigma^2} \|\y - \Bigh^{(l\minus1)} \x\|^2_2 + \lambda^{(l\minus1)} \|\x\|_1.
\end{aligned}
\end{equation}

Assuming the posterior at iteration $l$ is concentrated around its mode, i.e., $P(\x \mid \y ,\Bigh^{(l\minus1)},\lambda^{(l\minus1)};\sigma^2) \approx \delta(\x-\x^{(l)})$, we can approximate Eq.~(\ref{eq:qfunc}) by evaluating the complete-data log-likelihood at the posterior mode
\begin{equation}\label{eq:qfuncapprox}
\begin{aligned}
\mathcal{Q}(\theta|\theta^{(l-1)}) & \approx \log{P(\y , \x^{(l)}, \Bigh, \lambda; \sigma^2)} \\
 & \propto -\frac{1}{2\sigma^2}\|\y - \Bigh \x^{(l)}\|^2_2  - \lambda \|\x^{(l)}\|_1\\
& + C\log{P(\lambda)} + N_eC \log{\lambda}.
\end{aligned}
\end{equation}

\noindent Unless otherwise stated, the mode of the posterior is the sufficient statistics of the E-step, assuming the posterior concentrates around it. In Section~\ref{sec:crsae}, we argue that approximate EM gives a useful interpretation of AEs.

\vspace*{1mm}
\noindent \underline{\textbf{M-step}}: The M-step of the EM algorithm maximizes Eq.~(\ref{eq:qfuncapprox}) w.r.t. $\Bigh$ and $\lambda$. Hence, given $\x^{(l)}$, we update $\Bigh$ and $\lambda$ as follows
\vspace*{-2mm}
\begin{equation}\label{eq:Mstep1}
	\Bigh^{(l)} = \argmin_{\Bigh} \ \frac{1}{2} \|\y - \Bigh  \x^{(l)}\|^2_2
\end{equation}
\vspace*{-2mm}
\begin{equation}\label{eq:Mstep2}
	\lambda^{(l)} = \argmin_{\lambda} \ \lambda (\|\x^{(l)}\|_1 +  C\delta)  - (N_e + r - 1)C \log{\lambda}.
\end{equation}
\vspace*{-4mm}

\noindent Both objectives are convex w.r.t. to the parameter of interest, and the updates are available in closed-form. We also note that $\Bigh$ is a convolutional filter, and a constraint on the norm of the filters can be enforced similar to Eq.~(\ref{eq:CDL}).

The E-step parallels the CSC step in CDL. Eq.~(\ref{eq:Mstep1}) parallels the dictionary update step, while Eq.~(\ref{eq:Mstep2}) does not have an analogue in CDL. In principle, given $J$ independent examples $\y^j$, both the CSC step and the E-step are parallelizable. However, CDL and EM lack the infrastructure that would make it possible to solve the respective steps in parallel. The interpretation of CDL as EM motivates the AE we introduce in the next section. This architecture, called constrained recurrent sparse autoencoder (CRsAE), leverages the parallelism offered by GPUs to learn $\Bigh$ by gradient descent at a fraction of the time required by state-of-the-art CDL algorithms. Besides, CRsAE is able to learn $\lambda$, unlike existing CDL algorithms.

%%%%%%%%%%%%%%%%%%%%%%%%%%%%%%%%%%%%%%%%%%%%%%%%%%%%%%%%%%%%%%%%%%%%%%
\vspace*{-4mm}
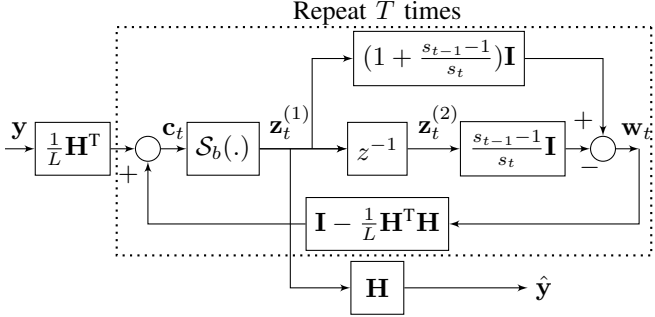
\begin{figure}[htb]
\begin{minipage}[b]{1.0\linewidth}
  \centering
\tikzstyle{block} = [draw, fill=none, rectangle, 
    minimum height=2em, minimum width=2em]
\tikzstyle{sum} = [draw, fill=none, circle, node distance=1cm]
\tikzstyle{input} = [coordinate]
\tikzstyle{output} = [coordinate]
\tikzstyle{pinstyle} = [pin edge={to-,thin,black}]
\begin{tikzpicture}[auto, node distance=2cm,>=latex']
	cloud/.style={
      draw=red,
      thick,
      ellipse,
      fill=none,
      minimum height=1em}
    % We start by placing the blocks
    \node [input, name=input] {};
    \node [block, node distance=0.9cm, right of=input] (HT) {$\frac{1}{L}\Bigh^\textrm{T}$};
    \node [sum, node distance=1cm, right of=HT] (sum) {};
    \node [block, right of=sum, node distance=1.cm] (shrinkage) {$\prox_{b}(.)$};
    % We draw an edge between the controller and system block to 
    \draw [->] (sum) -- node[name=u] {$\cvec_t$} (shrinkage);
    \node [block, node distance=2.05cm, right of=shrinkage] (delay) {$z^{-1}$};
    \node [block, node distance=1.8cm, right of=delay] (t) {$\frac{s_{t-1}-1}{s_t}\eye$};
    \node [block, node distance=1.0cm, below of=delay] (feedback) {$\eye-\frac{1}{L}\Bigh^\textrm{T}\Bigh$};
     
    \node [sum, node distance=1.2cm, right of=t] (sub) {};
    \node [output, node distance=0.48cm, right of=sub] (output) {};
  
    \node [rectangle, fill=none, node distance=1.81cm, above of=delay] (text) {Repeat $T$ times};
    % Once the nodes are placed, connecting them is easy. 
    \draw [draw,->] (input) -- node {$\y$} (HT);
    \draw [->] (HT) -- node {} (sum);
    \draw [->] (shrinkage) -- node [above,pos=0.35,name=Z] {$\z_t^{(1)}$}(delay);
    
    \draw [->] (shrinkage) -- node [below,name=temp,pos=0.6] {}(delay);

    \draw [->] (feedback) -| node[pos=0.90] {$+$} 
        node [near end] {} (sum);
    \draw [->] (delay) -- node[name=Z2, pos=0.6] {$\z_t^{(2)}$} (t);
    \node [block, node distance=0.85cm, above of=Z2] (forward) {$(1 + \frac{s_{t-1}-1}{s_t})\eye$};
    \draw [->] (temp) |- (forward);
    \draw [->] (forward) -| node[left,pos=0.90] {$\small{+}$} 
        node [near end] {} (sub);
	\draw [->] (t) -- node[pos=0.99, below] {$-$} (sub);	
	\draw[thick,dotted]     ($(forward.north east)+(+1.68,0.05)$) rectangle ($(HT.south east)+(+0.08,-1.06)$);
	\draw [->] (sub) -- node[name=wt,pos=0.90] {$\w_t$} (output);
	\draw [->] (output) |- (feedback);

	\node [block, node distance=0.85cm, below of=feedback] (H) {$\Bigh$};
	\draw [->] (Z) |- (H);
	\node[output, node distance=2cm, right of=H] (yhat) {};
	\draw [->] (H) -- node[right, pos=0.99] {$\hat \y$} (yhat);
\end{tikzpicture}
\end{minipage}
\vspace*{-7mm}
\caption{Block Diagram of CRsAE. Given a convolutional filter, the encoder is a recurrent network performing $T$ iterations of FISTA. The decoder applies the filters to the output of the encoder to reconstruct the input. The operator $z^{-1}$ refers to the delay in discrete-time. $\prox_b$ is a proximal operator (two-sided ReLU non-linearity). The bias $b = \frac{\lambda \sigma^2}{L}$, is a function of the regularization parameter $\lambda$, background noise $\sigma^2$, and the encoder step-size $L$.}
\label{fig:blockdiagram}
\vspace*{-6mm}
\end{figure}

\section{Proposed Network: Constrained Recurrent Sparse autoencoders}\label{sec:crsae}

\noindent When we first introduced CRsAE~\cite{TolooshamsBahareh2018SCDL}, it was as an interpretable AE for CDL. The original architecture, shown in Figure~\ref{fig:blockdiagram} and which assumes known $\lambda$, consists of an encoder, a decoder, and the norm-squared loss. It performs CDL by back-propagation through the AE to learn $\Bigh$. Here, we leverage the connection between CDL and EM to extend the architecture further in such a way that both $\Bigh$ and $\lambda$ can be learned.

%%%%%%%%%%%%%%%%%%%%%%%%%%%%%%%%%%

\subsection{CRsAE Architecture for EM-inspired Dictionary Learning}
\label{sec:CRSAE}
\begin{figure}[htb]
\begin{minipage}[b]{1.0\linewidth}
  \centering
\tikzstyle{block} = [draw, fill=none, rectangle, 
    minimum height=2em, minimum width=2em]
\tikzstyle{blueblock} = [draw, fill=none, rectangle, 
    minimum height=2em, minimum width=2em, color=blue]
 \tikzstyle{greenblock} = [draw, fill=none, rectangle, 
    minimum height=2em, minimum width=2em, color=orange]
\tikzstyle{sum} = [draw, fill=none, circle, node distance=1cm]
\tikzstyle{cir} = [draw, fill=none, circle, line width=1mm, minimum width=0.7cm, node distance=1cm]
\tikzstyle{loss} = [draw, fill=none, color=black, ellipse, line width=0.5mm, minimum width=0.7cm, node distance=1cm]
\tikzstyle{blueloss} = [draw, fill=none, color=black, ellipse, line width=0.5mm, minimum width=0.7cm, node distance=1cm, color=blue]
\tikzstyle{greenloss} = [draw, fill=none, color=black, ellipse, line width=0.5mm, minimum width=0.7cm, node distance=1cm, color=orange]
\tikzstyle{input} = [coordinate]
\tikzstyle{output} = [coordinate]
\tikzstyle{pinstyle} = [pin edge={to-,thin,black}]
\begin{tikzpicture}[auto, node distance=2cm,>=latex']
	cloud/.style={
      draw=red,
      thick,
      ellipse,
      fill=none,
      minimum height=1em}
    % We start by placing the blocks
    \node [input, name=input] {};
    \node [cir, node distance=1.cm, above of=input] (Y) {$\y$};
    \node [blueblock, above of=Y,  minimum width=1cm, node distance=1.0cm] (HT) {$\frac{1}{L}\Bigh^{\text{T}}$};
    \node [sum, node distance=0.9cm, above of=HT] (sum) {+};
    \node [greenblock, above of=sum,  minimum width=1cm, node distance=0.9cm] (relu) {$\prox_{b}(.)$};
    \node [cir, above of=relu, node distance=1.0cm] (xt) {$\x_t$};
    \node [cir, above of=xt, node distance=1.2cm] (xT) {$\x_T$};

    \node [output, node distance=0.9cm, above of=xT] (output) {};

    \node [blueblock, left of=output, node distance=0.8cm] (H) {$\Bigh$};
    \node [cir, left of=H, node distance=1.2cm] (Yhat) {$\hat \y$};
    \node [cir, left of=Yhat, node distance=1.2cm] (Y1) {$\y$};
    \node [blueblock, left of=relu, node distance=1.6cm] (f) {$g(.)$};
    
    \node [rectangle, fill=none,  node distance=0.6cm,  left of=Yhat] (middle) {};
    \node [blueloss, above of=middle, node distance=1.cm] (mse) {loss $\mathcal{L}_{\Bigh}$};
    
    \draw[thick, line width=2, black, ->]     ($(mse.south)+(+0,-1.2)$) -- ($(mse.south)+(-0,-6.82)$);
    \node [rectangle, fill=none,  node distance=3.9cm,  left of=xt] (back) {\footnotesize{Backprop for $\Bigh$}};
        \node [rectangle, fill=none,  node distance=0.4cm,  below of=back] (backD) {\footnotesize{(Dictionary update)}};
    \node [rectangle, fill=none,  node distance=0.4cm,  below of=backD] (backM) {\footnotesize{(M-step)}};

    \draw[thick,dotted]     ($(xT.north east)+(+0.3,-1)$) rectangle ($(sum.south west)+(-1.9,-1.23)$); 
    \node [rectangle, fill=none,  node distance=0.6cm,  above of=H] (text) {\footnotesize{Decoder}};
   \node [rectangle, fill=none,  node distance=1.8cm,  right of=relu] (encoder) {\footnotesize{Encoder}};
   \node [rectangle, fill=none,  node distance=0.4cm,  below of=encoder] (cscupdate) {\footnotesize{(CSC update)}};
   \node [rectangle, fill=none,  node distance=0.4cm,  below of=cscupdate] (Estep) {\footnotesize{(E-step)}};
    
    \draw[thick, line width=2, black, ->]     ($(Y.north east)+(0.60,0)$) -- ($(xT.south east)+(0.60,0)$);
  
    \node [greenloss, right of=mse, node distance=5.7cm] (lambdaloss) {loss $\mathcal{L}_{\lambda}$};
    \draw[thick, line width=2, black, ->]     ($(lambdaloss)+(+0,-1.0)$) -- ($(lambdaloss)+(-0,-7.2)$);
    \node [rectangle, fill=none,  node distance=2.1cm,  right of=xT] (back) {\footnotesize{Backprop for $\lambda$}};
    \node [rectangle, fill=none,  node distance=0.4cm,  below of=back] (backM) {\footnotesize{(M-step)}};

    % Once the nodes are placed, connecting them is easy. 
    \draw [->] (Y) -- node [name=m, midway, right] {} (HT);
    \draw [->] (HT) -- node {} (sum);
    \draw [->] (sum) -- node[] {} (relu);
    \draw [->] (relu) -- node[] {} (xt);
    \draw [->] (xt) -- node[name=loop, midway, pos=0.1, right] {} (xT);
    \draw [->] (xT) -- node[] {} (H);
    \draw [->] (H) -- node[name=z, pos=0.18] {} (Yhat);
    \draw [->] (loop) -| node[] {} (f);
    \draw [->] (f) |- node[] {} (sum);

    \node [rectangle, fill=none,  node distance=1.3cm,  below of=z] (text) {\footnotesize{Repeat $T$ times}};
    
    \draw [black, dashed, thick] (Yhat) --  (mse);
    \draw [black, dashed, thick] (Y1) --  (mse);
    
    \draw [black, dashed, thick] (xT) --  (lambdaloss);
         
\end{tikzpicture}
\end{minipage}
\vspace*{-7mm}
\caption{CRsAE architecture and two-stage training procedure. Given the input data $\y$, the encoder, which mimics the E-step of EM, outputs a sparse code $\x_T$ after $T$ iterations. Given $\x_T$, in a stage that mimics the first stage of the M-step, CRsAE updates $\Bigh$ by back-propagation through the reconstruction loss of the decoder. In the second stage, CRsAE back-propagates through the encoder to update $\lambda$ using a loss function motivated by Bayesian statistics. $g(\x_t) = (\eye - \frac{1}{L}\Bigh^{\text{T}}\Bigh)(\x_t + \frac{s_t - 1}{s_{t+1}}(\x_t - \x_{t-1}))$, and the bias $b = \frac{\lambda \sigma^2}{L}$.}
\label{fig:backprop}
\vspace*{-4mm}
\end{figure}
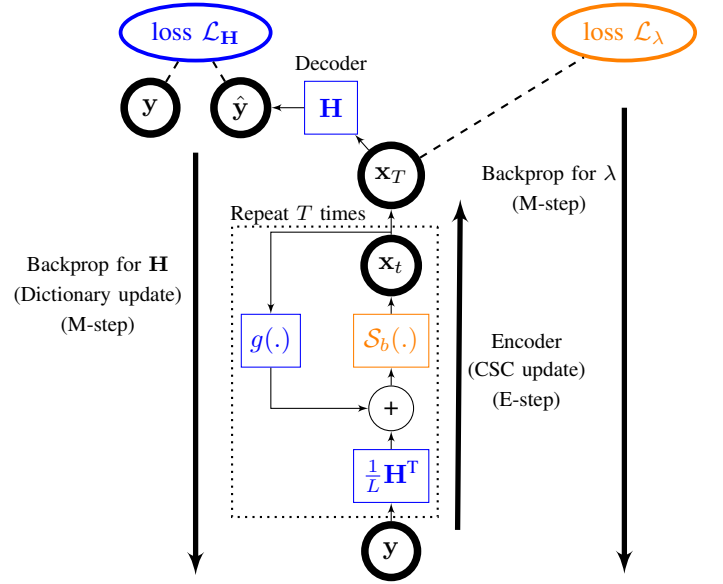

CRsAE consists of an encoder, a decoder and two loss functions. We learn the filters $\Bigh$ and the regularization parameter $\lambda$ through a two-stage back-propagation that uses one of the loss functions to update $\Bigh$ and the other to update $\lambda$. Figure~\ref{fig:backprop} shows the EM-inspired CRsAE architecture.

\subsubsection{Encoder, E-step and sparse coding}

Given the filters $\Bigh$ and the regularization parameter $\lambda$, the goal of the encoder is to map inputs $\y$ into sparse codes. The forward pass of the encoder obtains the sparse codes as the solution to the E-step (Eq.~(\ref{eq:Estep1})), which is equivalent to a CSC problem (Eq.~(\ref{eq:csc_update})). This optimization problem is given by
\begin{equation}\label{eq:encd}
\begin{aligned}
\min_{\x} \frac{1}{2\sigma^2}\|\y - \Bigh \x\|_2^2 +\lambda \|\x\|_1,
\end{aligned}
\end{equation}
\noindent where, for notational convenience, we ignore the indices $l$ and $j$ from Eqs.~(\ref{eq:Estep1}) and~(\ref{eq:csc_update}), respectively.

\noindent \underline{\textbf{CSC by ISTA}}: The ISTA algorithm~\cite{Daubechies2003AnIT} is a popular iterative method to solve Eq.~(\ref{eq:encd}). Let $\x_{t-1}$ represent the sparse code after $t$$-$$1$ iterations, ISTA updates $\x_{t-1}$ using the proximal gradient mapping
\begin{equation}
	\begin{aligned}
		\x_t = \prox_{\frac{\lambda \sigma^2}{L}}\left( \x_{t-1} + \frac{1}{L}\Bigh^T\left(\y - \Bigh \x_{t-1}\right) \right),
	\end{aligned}
\end{equation}
\noindent where the constant $L \geq \sigma_\text{max}(\Bigh^{\text{T}}\Bigh)$, and the operator, $\prox_{b}: \R^{N_e C} \to \R^{N_e C}$ is an element-wise two-sided ReLU
\begin{equation}\label{eq:shrinkage}
\begin{aligned}
(\prox_{b}(\z))_n & = \text{ReLU}(|z[n]|-b) \textrm{sgn}(z[n]) \\
	& = \text{ReLU}(z[n]-b) - \text{ReLU}(-z[n]-b),
\end{aligned}
\end{equation}

\noindent also known as shrinkage or soft-thresholding in the signal processing literature~\cite{Chen1998AtomicDB}. The function $\textrm{sgn}(\cdot)$ applies element-wise to its input, and equals $1$ if the entry is positive, $-1$ if it is negative, and $0$ otherwise. As detailed in~\cite{PapyanV2017CNNA}, the ReLU arises because of the regularization on the $\ell_1$-norm of $\x$ in Eq.~(\ref{eq:encd}). The ReLU is two-sided as the entries of $\x$ can be negative. When the entries cannot be negative, the operator becomes the ReLU nonlinearity commonly used in neural networks. By unfolding $T$ iterations of ISTA, we would obtain a neural network with ReLU nonlinearities that computes sparse codes given an input $\y$. We build the encoder of CRsAE by unfolding $T$ iterations of FISTA~\cite{beck2009fast}, fast ISTA, to solve Eq.~(\ref{eq:encd}) (or equivalently the E-step of Eq.~(\ref{eq:Estep1})). The sparsity of the encoder's output is a consequence of the depth $T$ of the encoder \emph{and} the shrinkage operator. For convolutional dictionaries, where the columns of $\Bigh$ are highly correlated, the encoder must be deep to result in a \emph{sparse} code at its output.

\noindent \underline{\textbf{Fast ISTA}}:  FISTA is an extension of ISTA that introduces a momentum term to accelerate convergence. Let $\x_{-1} = \x_0$, and let the ``state'' vector $\z_t = \begin{bmatrix}\z_{t}^{(1)}\  \z_{t}^{(2)}\end{bmatrix}^{\text{T}} = \begin{bmatrix}\x_{t}\ \x_{t-1}\end{bmatrix}^{\text{T}}$. Algorithm~\ref{algo:crsaeencoder} shows the encoder of CRsAE. It implements the FISTA algorithm. The constant $L$ is as defined previously, and $\x_t$ represents the sparse code after $t$ iterations of FISTA.

\noindent \underline{\textbf{Recurrent encoder by unfolding FISTA}}: Algorithm~\ref{algo:crsaeencoder} defines a recurrence relation $\z_t = f(\y,\z_{t-1},\Bigh)$ where $f(.)$ represents the operation in a single FISTA iteration. The encoder performs deep unfolding\cite{HersheyJohnR2014DUMI} and is built by unfolding this recurrent relation $T$ times, as depicted in the dashed box from Figure~\ref{fig:blockdiagram}. Each step in the recurrence employs linear operators $\Bigh$ and $\Bigh^{\text{T}}$ and the non-linear shrinkage operator $\prox_{b}$ (two-sided ReLU). In our convolutional setting, $\Bigh: \R^{N_eC} \to \R^{N}$ is a linear mapping from encoded sparse codes space to the data space performing a sum of convolutions as in Eq.~(\ref{eq:conv}). $\Bigh^{\textrm{T}}: \R^{N} \to \R^{N_eC}$ maps the input data into sparse code space while computing correlation between its input and filters $(\smlh_c)_{c=1}^C$. The output of the encoder is the code at the last iteration, namely $\x_T = \z_{T}^{(1)}$. Similar to the network in~\cite{Rolfe2013DiscriminativeRS}, all the time steps of this recurrent neural network share the same input $\y$, unlike classical recurrent architectures in which the input is fed in a sequence. This recurrent behaviour, shown as a for loop in Algorithm~\ref{algo:crsaeencoder}, allows us to produce approximately sparse codes, which are essential for dictionary learning~\cite{Agarwal2016LearningSU}.

\begin{algorithm}
\KwIn{$\y, \smlh, \lambda, \sigma, L \geq \sigma_\text{max}(\Bigh^{\text{T}}\Bigh)$}
\KwOut{$ \z_{T}^{(1)}$}
$\z_0 = \mathbf{0},s_0 = 0$\\
\For{$t =1$ to $T$}{
$s_t = \frac{1 + \sqrt{1+4s_{t-1}^2}}{2}$\\
$\w_t = \z_{t-1}^{(1)} + \frac{s_{t-1}-1}{s_t}\left(\z_{t-1}^{(1)}-\z_{t-1}^{(2)}\right) = \begin{bmatrix}\left(1 + \frac{s_{t-1}-1}{s_t}\right) \eye_{N_eC}| - \frac{s_{t-1}-1}{s_t} \eye_{N_eC}\end{bmatrix} \z_{t-1}$ \\
$\cvec_t = \w_t + \frac{1}{L} \Bigh^{\text{T}}(\y-\Bigh\w_t)$\\
$\z_t = \begin{bmatrix}\prox_{\frac{\lambda \sigma^2}{L}}(\cvec_t) \quad \z_{t-1}^{(1)}\end{bmatrix}^{\text{T}}$
}
\caption{$\text{CRsAE}_{\text{enc.}}(\y, \smlh, \lambda, \sigma, L)$: Encoder of CRsAE algorithm for producing sparse codes.}
\label{algo:crsaeencoder}
\end{algorithm}
\vspace{-2mm}

\vspace*{-4mm}
\begin{figure}[htb]
\begin{minipage}[b]{1.0\linewidth}
  \centering
\tikzstyle{block} = [draw, fill=none, rectangle, 
    minimum height=2em, minimum width=2em]
\tikzstyle{sum} = [draw, fill=none, circle, node distance=1cm]
\tikzstyle{input} = [coordinate]
\tikzstyle{output} = [coordinate]
\tikzstyle{pinstyle} = [pin edge={to-,thin,black}]
\begin{tikzpicture}[auto, node distance=2cm,>=latex']
	cloud/.style={
      draw=red,
      thick,
      ellipse,
      fill=none,
      minimum height=1em}
    % We start by placing the blocks
    \node [input, name=input] {};
    \node [rectangle, fill=none,  node distance=0.5cm,  above of=input] (text) {(a) ResNet};
    \node [block, node distance=1.cm, minimum width=1cm, below of=input] (H) {layer};
    \node [block, below of=H,  minimum width=1cm, node distance=2.cm] (HT) {layer};
    \node [sum, node distance=1.cm, below of=HT] (sum2) {+};
    \node [output, node distance=1.cm, below of=sum2] (output) {};
  
    % Once the nodes are placed, connecting them is easy. 
    \draw [->] (input) -- node[name=xtname, pos=0.3, right] {$\x_{t}$} (H);
    \draw [->] (input) -- node[name=xt, midway, left] {} (H);
    \draw [->] (H) -- node [name=m, midway, right] {ReLU} (HT);
    \draw [->] (HT) -- node {} (sum2);
    \draw [->] (sum2) -- node[name=relu] {ReLU} (output);
    \draw[->] (xt) to [out=1,in=1] (sum2); 
       
    \node [rectangle, fill=none,  node distance=1.7cm,  left of=m] (F) {$\mathcal{F}(\x_t)$};
    \node [rectangle, fill=none,  node distance=1.40cm,  left of=relu] (F) {$\mathcal{F}(\x_t)$$+$$\x_t$};

%%%%%%%%%%%%%%%%%%%%%%%%%%%%%%%%%%%%%%%%%%     
%%%%%%%%%%%%%%%%%%%%%%%%%%%%%%%%%%%%%%%%%%     

    \node [input, name=input_1, node distance=2.1 cm, right of=input] {};
    \node [rectangle, fill=none,  node distance=0.5cm,  above of=input_1] (text) {(b) ISTA};
    \node [block, node distance=1.cm, minimum width=1cm, below of=input_1] (H_1) {$-\Bigh$};
    \node [sum, node distance=1cm, below of=H_1] (sum1_1) {+};
    \node [block, below of=sum1_1,  minimum width=1cm, node distance=1.cm] (HT_1) {$\frac{1}{L}\Bigh^{\text{T}}$};
    \node [sum, node distance=1.cm, below of=HT_1] (sum2_1) {+};
    \node [output, node distance=1.cm, below of=sum2_1] (output_1) {};
      
    % Once the nodes are placed, connecting them is easy. 
    \draw [->] (input_1) -- node[name=xt_1name, pos=0.3, right] {$\x_{t}$} (H_1);
    \draw [->] (input_1) -- node[name=xt_1, midway, left] {} (H_1);
    \draw [->] (H_1) -- node {} (sum1_1);
    \draw [->] (sum1_1) -- node {} (HT_1);
    \draw [->] (HT_1) -- node {} (sum2_1);
    \draw [->] (sum2_1) -- node[name=relu_1] {ReLU} (output_1);
    \draw[->] (xt_1) to [out=1,in=1] (sum2_1); 
        
     \node [input, node distance=0.7cm,  right of=sum1_1] (y) {};
     \draw[->] (y) -- node[above] {$\y$} (sum1_1);
     
%%%%%%%%%%%%%%%%%%%%%%%%%%%%%%%%%%%%%%%%%%     
%%%%%%%%%%%%%%%%%%%%%%%%%%%%%%%%%%%%%%%%%%     

    \node [input, name=input_2, node distance=2.1 cm, right of=input_1] {};
    \node [rectangle, fill=none,  node distance=0.5cm,  above of=input_2] (text) {(c) FISTA};
    \node [block, node distance=1.cm, minimum width=1cm, below of=input_2] (H_2) {$-\Bigh$};
    \node [sum, node distance=1cm, below of=H_2] (sum1_2) {+};
    \node [block, below of=sum1_2,  minimum width=1cm, node distance=1.cm] (HT_2) {$\frac{1}{L}\Bigh^{\text{T}}$};
    \node [sum, node distance=1.cm, below of=HT_2] (sum2_2) {+};
    \node [output, node distance=1.cm, below of=sum2_2] (output_2) {};
      
    % Once the nodes are placed, connecting them is easy.
    \draw [->] (input_2) -- node[name=xt_2, pos=0.3, right] {$\w_{t}$$=$$f(\x_{t-1}, \x_{t-2})$} (H_2); 
    \draw [->] (input_2) -- node[name=xt_2, midway, left] {} (H_2);
    \draw [->] (H_2) -- node {} (sum1_2);
    \draw [->] (sum1_2) -- node {} (HT_2);
    \draw [->] (HT_2) -- node {} (sum2_2);
    \draw [->] (sum2_2) -- node[name=relu_2] {ReLU} (output_2);
    \draw[->] (xt_2) to [out=1,in=1] (sum2_2); 
    
     \node [input, node distance=0.7cm,  right of=sum1_2] (y_2) {};
     \draw[->] (y_2) -- node[above] {$\y$} (sum1_2);  
         
\end{tikzpicture}
\end{minipage}
\vspace*{-7mm}
\caption{ISTA and FISTA as ResNets. The building blocks of a ResNet, ISTA, and FISTA are shown in (a), (b), and (c), respectively. FISTA performs a residual learning similar to ISTA on a vector $\w_{t}$ which is a function of the sparse code $\x$ in unfolded layers $t$-$1$ and $t$-$2$ given by: $f(\x_{t-1}, \x_{t-2}) = \x_{t-1} + \frac{s_{t-1} - 1}{s_t} (\x_{t-1} - \x_{t-2})$.}
\label{fig:resnet}
\vspace*{-4mm}
\end{figure}
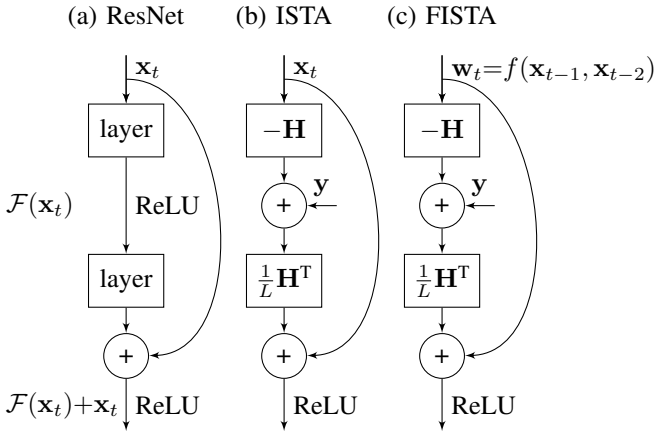

\noindent \underline{\textbf{Encoder and ResNet}}: The encoder in CRsAE resembles a deep residual network~\cite{HeKaiming2015DRLf}. In fact, when unfolded, both ISTA and FISTA can be interpreted as variants of deep residual networks as also noted in~\cite{mardani2018}. Figure~\ref{fig:resnet}(a) shows one block of a residual network. As depicted in the figure, ISTA (Figure~\ref{fig:resnet}(b)) and FISTA (Figure~\ref{fig:resnet}(c)) each imposes a specific structure on the form of the residual mapping  $\mathcal{F}(\x)$.

\subsubsection{Decoder, M-step and loss functions}
The M-step of EM from Section~\ref{sec:EM} dictates the form of the decoder, and the loss functions to update the trainable parameters. Specifically, the M-step suggests a two-stage back-propagation procedure (Figure~\ref{fig:backprop}) for updating $\Bigh$ and $\lambda$ that is the subject of the next sub-section. Here, we argue that the two-stage training is the natural consequence of using Eqs.~(\ref{eq:Mstep1}) and~(\ref{eq:Mstep2}) to augment the encoder described previously.

\vspace*{1mm}
\noindent \underline{\textbf{M-step/$\Bigh$ update}}: The objective from Eq.~(\ref{eq:Mstep1}) motivates a linear decoder, which applies $\Bigh$ to the output of the encoder to reconstruct $\y$, and the norm-squared as the loss to apply to the output of the decoder. The weights of the encoder and decoder are tied to each other. Hence, one branch of the AE, the one that will be used to update $\Bigh$, is constructed as
\vspace*{-1.0mm}
\begin{equation}\label{eq:ae}
\text{CRsAE}(\y, \smlh, \lambda, \sigma, L) = \Bigh\ \text{CRsAE}_{\text{enc.}}(\y, \smlh, \lambda, \sigma, L).
\end{equation}
\vspace*{-4.0mm}

\noindent In Figure~\ref{fig:blockdiagram}, we refer to $\z_T^{(1)}$ as the sparse code (output of the encoder). Constraining the encoder and decoder makes the filters interpretable in terms of the model in Eq.~(\ref{eq:DTconv}) and also the dictionary learning optimization problem in Eq.~(\ref{eq:CDL}). Moreover, it reduces the number of trainable parameters by a factor of $\frac{1}{3}$ compared to the AE in~\cite{SreterHillel2018LCSC}. Eq.~(\ref{eq:Mstep1}) suggests the following loss function, similar to~\cite{rethinking,Makhzani2013kSparseA}, to learn $\Bigh$:
\begin{equation}\label{eq:loss_H}
\mathcal{L}_H({\y, \smlh, \lambda, \sigma, L)} =  \half \|\y - \underbrace{\text{CRsAE}(\y, \smlh, \lambda, \sigma, L)}_{\hat 
\y}\|_2^2.
\end{equation}
\vspace*{-5.0mm}

\begin{table*}[!ht] 
\caption{Comparison of classical dictionary learning, EM-based dictionary learning, and CRsAE.}
\label{tab:methods}
\vspace*{-2mm}
  \centering
  \begin{tabular}{p{1cm}p{5.8cm}p{4.9cm}p{4.8cm}}
         & Sparse Coding ($\x$) & Dictionary Learning ($\Bigh$) & Regularization ($\lambda$) \\ \midrule
    Dictionary learning & 
    $\begin{aligned} \x^{(l)} = \argmin_{\x}\ \frac{1}{2\sigma^2}\|\y - \Bigh^{(l\minus1)}  \x\|_2^2 +\lambda \|\x\|_1\end{aligned}$ &
    $\begin{aligned}\label{eq:dict}
\Bigh^{(l)} = \argmin_{\{\smlh_c\}_{c=1}^C}\ &\frac{1}{2} \|\y-\Bigh  \x^{(l)} \|_2^2\\ \text{ s.t. } &\|\smlh_c\|_2 = 1
\end{aligned}$
 & Hyperparameter\\  \midrule
    EM &
    $\begin{aligned} \x^{(l)} = \argmax_{\x}\ \log{P(\x \mid \y,\Bigh^{(l\minus1)},\lambda^{(l\minus1)},\sigma^2}) \end{aligned}$ &
    $\begin{aligned}\Bigh^{(l)} = \argmin_{\{\smlh_c\}_{c=1}^C} \ &\frac{1}{2} \|\y - \Bigh  \x^{(l)}\|^2_2\\ \text{ s.t. } &\|\smlh_c\|_2 = 1 \end{aligned}$
    &
    $\begin{aligned} \lambda^{(l)} = &\argmin_{\lambda} \ \lambda (\|\x^{(l)}\|_1 +  C\delta)\\ &- (N_e + r - 1)C \log{\lambda}  \end{aligned}$ \\ \midrule
    CRsAE & $\text{CRsAE}_{\text{enc.}}(\y, \smlh^{(l\minus1)}, \lambda^{(l\minus1)}, \sigma, L)$ &
    $\smlh^{(l)} $\minus$= \eta_{\smlh}  \nabla_{\smlh} \mathcal{L}_H(\y, \smlh^{(l\minus1)}, \lambda^{(l\minus1)}, \sigma, L)$  &
    $\lambda^{(l)} $\minus$=\eta_{\lambda} \nabla_{\lambda} \mathcal{L}_{\lambda}({\y, \smlh^{(l\minus1)}, \lambda^{(l\minus1)}, \sigma, L)}$\\ \bottomrule
   \\
  \end{tabular}
\vspace*{-5mm}
\end{table*}

\noindent \underline{\textbf{M-step/$\lambda$ update}}: The objective function from Eq.~(\ref{eq:Mstep2}) suggests that we use the loss function from Eq.~(\ref{eq:Mstep2}) to update $\lambda$. Therefore, we use the following to learn $\lambda$:
\begin{equation}\label{eq:loss_lambda}
\begin{aligned}
\mathcal{L}_{\lambda}({\y, \smlh, \lambda, \sigma, L)} = &\lambda (\underbrace{\|\text{CRsAE}_{\text{enc.}}(\y, \smlh, \lambda, \sigma, L)\|_1}_{\|\z_T^{(1)}\|_1} + C\delta)\\
- &(N_e + r-1)C \log{\lambda} 
\end{aligned}
\end{equation}

\noindent Together, Algorithm~\ref{algo:crsaeencoder},~Eqs.~(\ref{eq:ae}),~(\ref{eq:loss_H}) and~(\ref{eq:loss_lambda}) fully specify the CRsAE architecture and the losses we use to train it. The structure of CRsAE resembles that of a variational autoencoder (VAE)~\cite{KingmaW13,chen2016variational}. VAEs are probabilistic AEs. The output of the encoder from a VAE is an approximate sample from the posterior distribution of the latent (e.g., $\x$) given the data. Section~\ref{sec:CONC} discusses this connection and what a CRsAE-like VAE would resemble.

\vspace*{-4.0mm}
\subsection{Back-propagation}\label{sec:backprop}

\noindent We train CRsAE by a two-stage back-propagation procedure. Figure~\ref{fig:backprop} illustrates the steps involved in this procedure. Minimizing the losses in this procedure corresponds to maximizing the expectation of the log complete-data likelihood using the approximate sufficient statistics, i.e., the MAP estimate of the codes, implicitly computed in the forward pass of CRsAE. 
 
In the first stage, we perform back-propagation through the AE using the loss function from Eq.~(\ref{eq:loss_H}). Algorithm~\ref{algo:crsaebpropdict} is the back-propagation algorithm for computing the gradient of the loss function in Eq.~(\ref{eq:loss_H}) w.r.t. $\Bigh$. The output of this algorithm is the gradient of the loss function in Eq.~(\ref{eq:loss_H}), denoted by $\delta \cdot$~\cite{gregor2010learning}, w.r.t. the set of filters $\{\smlh_c\}_{c=1}^C$ shared across all layers. The number of trainable parameters is only $K \times C$ and is equal to the number of filter parameters in one layer of the network.

%%%%%%%%%%%%%%%%%%%%%%%%%%%%%%%%%%
\vspace*{-1.0mm}
\begin{algorithm}
\KwIn{$\y,\lambda,\sigma,L,\smlh$, $\{s_t\}_{t=1}^T, \{\w_t\}_{t=1}^T, \{\cvec_t\}_{t=1}^T, \{\z_t\}_{t=1}^T$.}
\KwOut{$\delta \smlh$}
$\delta \hat \y = \hat \y - \y, \delta \smlh = \mathbf{0}_K$\\
$\delta \cvec_{T+1} = \frac{\partial \hat \y}{\partial \cvec_{T+1}} \delta \hat \y $\\
$\delta \smlh = \delta \smlh +  \frac{\partial \cvec_{T+1}}{\partial \smlh} \delta \cvec_{T+1}$\\
$\delta \z_{T} = \frac{\partial \cvec_{T+1}}{\partial \z_{T}} \delta \cvec_{T+1}$\\
%%%%%%
\For{$t =T$ to $1$}{
$\delta \cvec_t = \frac{\partial \z_t}{\partial \cvec_t} \delta \z_t $\\
$\delta \smlh = \delta \smlh +  \frac{\partial \cvec_t}{\partial \smlh} \delta \cvec_t$\\
$\delta \z_{t-1} = \frac{\partial \cvec_t}{\partial \z_{t-1}} \delta \cvec_t$
}
\caption{$\nabla_{\smlh} \mathcal{L}_H(\y, \smlh, \lambda, \sigma, L)$: Back-propagation of CRsAE for filters.}
\label{algo:crsaebpropdict}
\end{algorithm}

\begin{algorithm}
\KwIn{$\y,\lambda,\sigma, L,\smlh$, $\{s_t\}_{t=1}^T, \{\w_t\}_{t=1}^T, \{\cvec_t\}_{t=1}^T, \{\z_t\}_{t=1}^T$}
\KwOut{$\delta \lambda$}
$(\frac{\partial  \|\z_T^{(1)}\|_1}{\partial \z_{T}})_i = \begin{cases}
	\text{sign}(\z_{T}[i])\ \text{if}\ \z_{T}[i] \neq 0 \\
	0\quad \text{,otherwise}
\end{cases}$\\
$\delta \lambda =  \|\z_T^{(1)}\|_1 + C\delta - \frac{(N_e+ r - 1)C}{\lambda}  +  \lambda \frac{\partial \z_T}{\partial \lambda} \frac{\partial \|\z_T^{(1)}\|_1}{\partial \z_{T}}$\\
\For{$t =T$ to $1$}{
$ \frac{\partial \z_t}{\partial \z_{t-1}} =  \frac{\partial \z_t}{\partial \cvec_{t}}  \frac{\partial \cvec_t}{\partial \z_{t-1}}$\\
$\frac{\partial \|\z_T^{(1)}\|_1}{\partial \z_{t-1}} = \frac{\partial \z_t}{\partial \z_{t-1}} \frac{\partial  \|\z_T^{(1)}\|_1}{\partial \z_t}$\\
$\delta \lambda = \delta \lambda +  \lambda \frac{\partial \z_{t-1}}{\partial \lambda} \frac{\partial \|\z_T^{(1)}\|_1}{\partial \z_{t-1}}$}
\caption{$\nabla_{\lambda} \mathcal{L}_{\lambda}({\y, \smlh, \lambda,\sigma, L)}$: Back-propagation of CRsAE for regularization parameter.}
\label{algo:crsaebproplambda}

\end{algorithm}

In the second stage, we perform back-propagation through the encoder using the loss function from Eq.~(\ref{eq:loss_lambda}). This latter stage is unique to our approach and does not have an equivalent either in  dictionary learning, where $\lambda$ is typically not trained, or in deep learning. The back-propagation algorithm for computing the gradient of the loss function in Eq.~(\ref{eq:loss_lambda}) w.r.t. $\lambda$ is given in Algorithm~\ref{algo:crsaebproplambda}. The term $- (N_e + r - 1)C \log{\lambda}$ plays an important role in the successful estimation of $\lambda$ as it prevents its convergence to zero during learning. In other words, training $\lambda$ na\"ively through back-propagation by only minimizing the first component of Eq.~(\ref{eq:loss_lambda}) may result in convergence of $\lambda$ to zero. If $\lambda$ were to converge to zero, the encoder would fail to produce sparse codes. This implies that, in the absence of the second term from Eq.~(\ref{eq:loss_lambda}), if the AE were fed simulated data generated by known filters $\Bigh$ and sparse codes, the overall two-stage procedure would fail to learn $\Bigh$ because the success of dictionary learning relies on the encoder producing sparse codes~\cite{Agarwal2016LearningSU}.

The derivations for Algorithms~\ref{algo:crsaebpropdict} and~\ref{algo:crsaebproplambda} are provided in detail in the Appendix. Algorithm~\ref{algo:EMtraining} summarizes the two-stage back-propagation procedure, which, to our knowledge, is novel. Any gradient-based method can be used for training. For simplicity, Algorithm~\ref{algo:EMtraining} uses gradient descent to learn the parameters. Table~\ref{tab:methods} summarizes the parallel between the steps in classical CDL, EM-inspired CDL, and CRsAE.

\vspace{-4mm}
\begin{algorithm}%[htb]
\KwIn{$\y, \smlh^{\text{init}}, \lambda^{\text{init}}, \sigma, L \geq \sigma_\text{max}(\Bigh_{\text{init}}^{\text{T}}\Bigh_{\text{init}}), \eta_{\smlh}, \eta_{\lambda}, I $}
\KwOut{$\smlh^{(I)},\lambda^{(I)}$}
$\smlh^{(0)} = \smlh^{\text{init}}$\\
$\lambda^{(0)} = \lambda^{\text{init}}$\\
\For{$l$ = $1$ to $I$}{
$\smlh^{(l)} $\minus$= \eta_{\smlh}  \nabla_{\smlh} \mathcal{L}_H(\y, \smlh^{(l\minus1)}, \lambda^{(l\minus1)}, \sigma, L)$ \\
$\lambda^{(l)} $\minus$= \eta_{\lambda} \nabla_{\lambda} \mathcal{L}_{\lambda}({\y, \smlh^{(l\minus1)}, \lambda^{(l\minus1)}, \sigma, L)} $\\
$\|\smlh_c^{(l)}\|_2 = 1$
}
\caption{Training CRsAE.}
\label{algo:EMtraining}
\end{algorithm}
%%%%%%%%%%%%%%%%%%%%%%%%%%%%%%%%%%%%%%%%%%%%%%%%%%%%%%%%%%%%%%%%%%%%%%
\vspace{-10mm}
\subsection{Training}\label{sec:training}

We trained CRsAE through back-propagation on a GPU (NVIDIA Tesla V100) provided by AWS, using the ADAM optimizer~\cite{Kingma2014AdamAM}. When a validation set is available, we pick the parameters that minimize the validation loss over all epochs.

\noindent \underline{\textbf{Gamma hyper-prior parameters }{($\delta$, r)}}: Given a dense dictionary, a suggested value for $\lambda$ motivated by theory~\cite{Chen1998AtomicDB} is $\frac{\sqrt{2\log( C\times N_e)}}{\sigma}$. We set the parameters of the gamma prior on $\lambda$ such that the prior is centered around the suggested estimate. In CDL, we do not have access to the true filters, which results in an additional source of noise: $\y = \Bigh_{0} \x + \underbrace{(\Bigh-\Bigh_{0}) \x + \smlv}_{\smlv_0}$ where $\smlv_0$ contains observation noise and noise from the lack of knowledge of $\Bigh$. Hence, it is reasonable to expect $\lambda$ to deviate from the suggested estimate during training. To allow for this, we chose the $\delta$ such that the distribution is wide enough for $\lambda$ to deviate from its mean during training while maintaining stability by avoiding the convergence of $\lambda$ to 0 or $\infty$. For the simulated data from Section~\ref{sec:sim}, $\sigma$ is known. For the real data from Section~\ref{sec:real}, we can estimate $\sigma$ from ``silent'' periods in the signal.

\vspace*{1mm}
\noindent \underline{\textbf{Encoder hyperparameters ($L$, $T$)}}: We choose $L$ to be greater than $\sigma_{\text{max}}(\Bigh^\textrm{T}\Bigh)$~\cite{beck2009fast}. For neural data, we can estimate $L$ from an existing collection of action potentials. The value of $T$ does not affect the trainable parameters as the encoder is a recurrent network with shared parameters. However, for the encoder to produce sparse codes, $T$ should be large.

\vspace*{1mm}
\noindent \underline{\textbf{Optimizer hyperparameters {($\eta_{\smlh}$, $\eta_{\lambda}$, $B$)}}}: The learning rates ($\eta_{\smlh}$ for $\smlh$ and $\eta_{\lambda}$ for $\lambda$) of the optimizer depend on the smoothness of the loss function, which is a function of the model and dataset. For the filters, we found an optimal $\eta_{\smlh}$ range by varying it from $10^{-5}$ to $10^{-1}$ while monitoring the validation loss. As suggested in~\cite{Smith2017CyclicalLR}, in our 1D experiments, we picked the optimal $\eta_{\smlh}$ as the one that leads to the sharpest drop in the validation loss. We tuned the $\eta_{\lambda}$ to a value in the range between $0.1$ and $5$. We chose $\eta_{\lambda}$ to be large so that the effective learning rate for the bias $b = \lambda \sigma^2/L$ is appropriate for training, i.e., not too small. In addition, choosing a larger $\eta_{\lambda}$ compared to $\eta_{\smlh}$ allows the value of $\lambda$ to stabilize faster than $\Bigh$ within each epoch. As discussed in Section~\ref{sec:EXP}, we found that this behavior was crucial in allowing the training to converge to the ground-truth filters $\Bigh$.

The batch size $B$ corresponds to the number of examples used in every gradient update step of back-propagation. For the time-series dataset consisting of recordings of extracellular voltage from neurons, we found that it was important to relate $B$ to the expected number of spikes in each batch.  We can estimate the expected number of spikes in a window, given its length $N$ and the firing rate of neurons. In turn, this calculation yields the expected number of spikes in each batch. In our experiments, we chose $B$ so as to have enough spikes in each mini-batch but still have a large number of batches to take advantage of stochasticity during training.

\vspace*{1mm}
\noindent \underline{\textbf{Augmentation}}: We used two forms of augmentation for 1D experiments: flipping and circular rotation, which do not change the generative model (Eq.~(\ref{eq:DTconv})), but are useful as the training is done through mini-batch gradient descent on noisy data~\cite{KrizhevskyAlex2017Icwd}. In our model, augmentation by flipping changes the sign of the sparse codes and does not increase the complexity of the model, as the encoder uses two-sided shrinkage (Eq.~(\ref{eq:shrinkage})), i.e., as opposed to ReLU~\cite{GlorotXavier2011DSRN}. In augmentation by circular rotation, the sparse code positions are shifted. This augmentation is done by circularly rotating each example by an integer delay randomly generated between $1$ to $N$.

%%%%%%%%%%%%%%%%%%%%%%%%%%%%%%%%%%

\begin{table} 
\caption{Details of Datasets and Training Parameters for Experiments on 1D Neural Data.}
\label{tab:dataset1d}
\vspace*{-2mm}
  \centering
  \setlength\tabcolsep{5pt}
  \begin{tabular}{ccc}
           & Simulated & Real   \\ \midrule
    Length of data $T_0$ [min] &  $17$ & $4$  \\  \midrule
    Firing rate of each neuron [Hz] & $30$ & - \\ \midrule
    Sampling frequency $f_s$ [Hz] & $10{,}000$ & $10{,}000$  \\ \midrule
    Sparseness of code $\|\x^j\|_0$ & $3\times$$4$$=$$12$ & -  \\ \midrule
    $\sigma$ & set based on SNR& 0.03  \\  \midrule
    \# filters $C$ & $4$  & $2$ \\  \midrule
    Filter size $K$  & $18$   & $35$   \\  \midrule
    Length of each example $N$ & $1{,}000$   & $60{,}000$ \\  \midrule
    \# examples & $10{,}100$ & $24$ \\  \midrule
    \# training examples & $9{,}000$ & $63$ augmented  \\ \midrule
    \# validation examples & $1{,}000$ & $3$  \\ \midrule
    \# testing examples  & $100$ & -  \\ \midrule
    \# trainable parameters & $(18\times$$4)$$+ 1$$=$$73$ & $(35\times$$2)$$+ 1$$=$$71$  \\  \midrule  
     FISTA iterations $T$ &  $180$ & $600$\\  \midrule
     Batch size $B$ & $1024$ & $4$ \\ \midrule
     $L$ & $13.5$ & $15$  \\ \midrule
    $\lambda_{\text{init}}$ & $\frac{\sqrt{2\log( C\times N_e)}}{\sigma}$ & $\frac{\sqrt{2\log( C\times N_e)}}{\sigma}$ \\ \midrule
    $\delta$  & $50$ & $1{,}000$  \\ \midrule
    $r$ & $\delta \lambda_{\text{init}}$ & $\delta \lambda_{\text{init}}$\\ \midrule
    ${\smlh}^{\text{init}}$ & - & k-means \\ \midrule
    $\text{err}(\smlh_c,{\smlh}^{\text{init}}_c)$ & $-3$ to $-4$ & - \\ \midrule
  $\eta_{\smlh}$ & $10^{-5}$ to $10^{-1}$ & $10^{-5}$ to $10^{-1}$  \\  \midrule
  $\eta_{\lambda}$ & $1$ to $5$& $1$ to $5$ \\  \bottomrule
  \end{tabular}
\vspace*{-5mm}
\end{table}

\vspace{-3mm}
\section{Experiments}
\label{sec:EXP}

\noindent In this section, we apply CRsAE\footnote{https://github.com/ds2p/crsae} to two different 1D datasets and one two-dimensional (2D) dataset. The first 1D dataset consists of simulated recordings of extracellular voltage from neurons with known ground-truth filters. We use this example to compare the ability of CRsAE to learn, in an unsupervised manner, the ground-truth filters to the AE called learned convolutional sparse coding (LCSC) from~\cite{SreterHillel2018LCSC}. The encoder of LCSC uses $3$ ISTA iterations. Unlike in CRsAE, the decoder is \emph{unconstrained}, i.e., it is not tied to the encoder. In LCSC, each filter is associated with its own regularization parameter. The overall objective of LCSC is to minimize the least-squares (LS) reconstruction loss from input to output. We also use the simulated example to compare the computational efficiency of CRsAE to the state-of-art optimization-based CDL algorithm from~\cite{garcia-2018-convolutional}, which is implemented in the Sporco library~\cite{wohlberg-2017-sporco} (we refer to this method by Sporco). Sporco solves the CSC update through ADMM~\cite{wohlberg-2014-efficient}, and solves the dictionary update by an accelerated proximal gradient method.

The second 1D dataset consists of recordings of extracellular voltage from neurons in the Hippocampus of a rat~\cite{HarrisKd2000Aots}. We use this dataset to compare CRsAE to continuous basis pursuit (CBP)~\cite{ekanadham2014unified}, the state-of-the-art optimization-based algorithm for spike sorting, which is the process of identifying the location of action potentials (sparse codes) and their corresponding neurons in extracellular recordings.

The third dataset is a dataset of natural images. We use it to compare CRsAE to LCSC on an image denoising task. In this task, we study the effect of encoder depth, weight sharing between encoder and decoder, as well as the method for training the regularization parameter, on denoising performance.

\vspace*{-4mm}
\subsection{Simulated Data}\label{sec:sim}

In this experiment, we demonstrate the ability of CRsAE to recover, in an unsupervised setting, filters that appear in a 1D dataset. We use the following standard error~\cite{Agarwal2016LearningSU}
\begin{equation}\label{eq:distance_error}
\text{err}(\smlh_c,\hat{\smlh}_c) = 10 \log{\left(\sqrt{1 - \frac{\ip{\smlh_c}{\hat{\smlh}_c}^2}{\|{\smlh}_c\|_2^2\|\hat{\smlh}_c\|_2^2}}\right)},
\end{equation}
where $\hat{\smlh}_c$ is one of the learned filters, to compare the learned filters to the true ones. This error ranges from $0$ to $-\infty$. The closer the learned filter to the true one, the smaller $\text{err}(\smlh_c,\hat{\smlh}_c)$. We characterize the sensitivity of CRsAE to noise by computing this error for a range of SNR values.

\noindent \underline{\textbf{Simulated extra-cellular data:}} We simulated a recording (Eq.~(\ref{eq:CTconv})) of length approximately $T_0 = 17$ minutes consisting of the sum electrical-voltage activity from $C = 4$ neurons, each with an average firing rate of $30$ Hz. The unit of the recording was in mV. Consistent with the biophysics of neurons~\cite{lewicki1998review}, we picked filters (action potentials) of length $1.8$ ms.  We chose the amplitudes $\{x_{c,i}\}_{i=1}^{N_c}$ and times of occurrence $\{\tau_{c,i}\}_{i=1}^{N_c}$ of the filters independently for each neuron, respectively, according to a $\mathcal{N}(180,30)$ distribution and a Poisson process with rate $30$ Hz. The cross-correlation between the filters ranged from $0.5$ to $0.95$. 

We divided the voltage recording into windows of length $100$ ms, resulting in a total number $J = 10{,}100$ windows (examples). Assuming a sampling rate of $f_s = 10$ kHz, the length of each window was $N = 1{,}000$, and each filter was $K = 18$ samples long. Therefore, $\y^j \in \R^{1,000}$ and $\smlh_c \in \R^{18}$. For each example $j$, each of the vectors $\{\x_c^j\}_{c=1}^C$ was $3$-sparse. Because of the refractory period of neurons, whose length is on the same order as the filter length, occurrences of filters from a given neuron cannot overlap within the signal. This implies that, in each sparse code $\x_c^j$, the non-zero entries were at least $K$ samples apart. Filters from different neurons were allowed to overlap~\cite{lewicki1998review}. We added Gaussian noise $\smlv^j \sim \mathcal{N}(\textbf{0},\sigma^2 \eye)$ to each example $\y^j$, where the noise variance $\sigma^2$ was set according to a specified SNR. Finally, as is standard practice~\cite{LecunY1998Eb}, we normalized the dataset by its maximum absolute value. Note that this scaling did not affect the generative model, as all the examples were re-scaled by the same constant. The first column of Table~\ref{tab:dataset1d} summarizes the data and training parameters of the simulation. The depth of the network is $T+1 = 181$ layers, and the number of channels at every layer is $C=4$, the number of neurons. The width of each channel is $N-K+1 = 983$.

\begin{figure}[htb]
\begin{minipage}[b]{1.0\linewidth}
  \centering
  \centerline{\includegraphics[width=8.0cm]{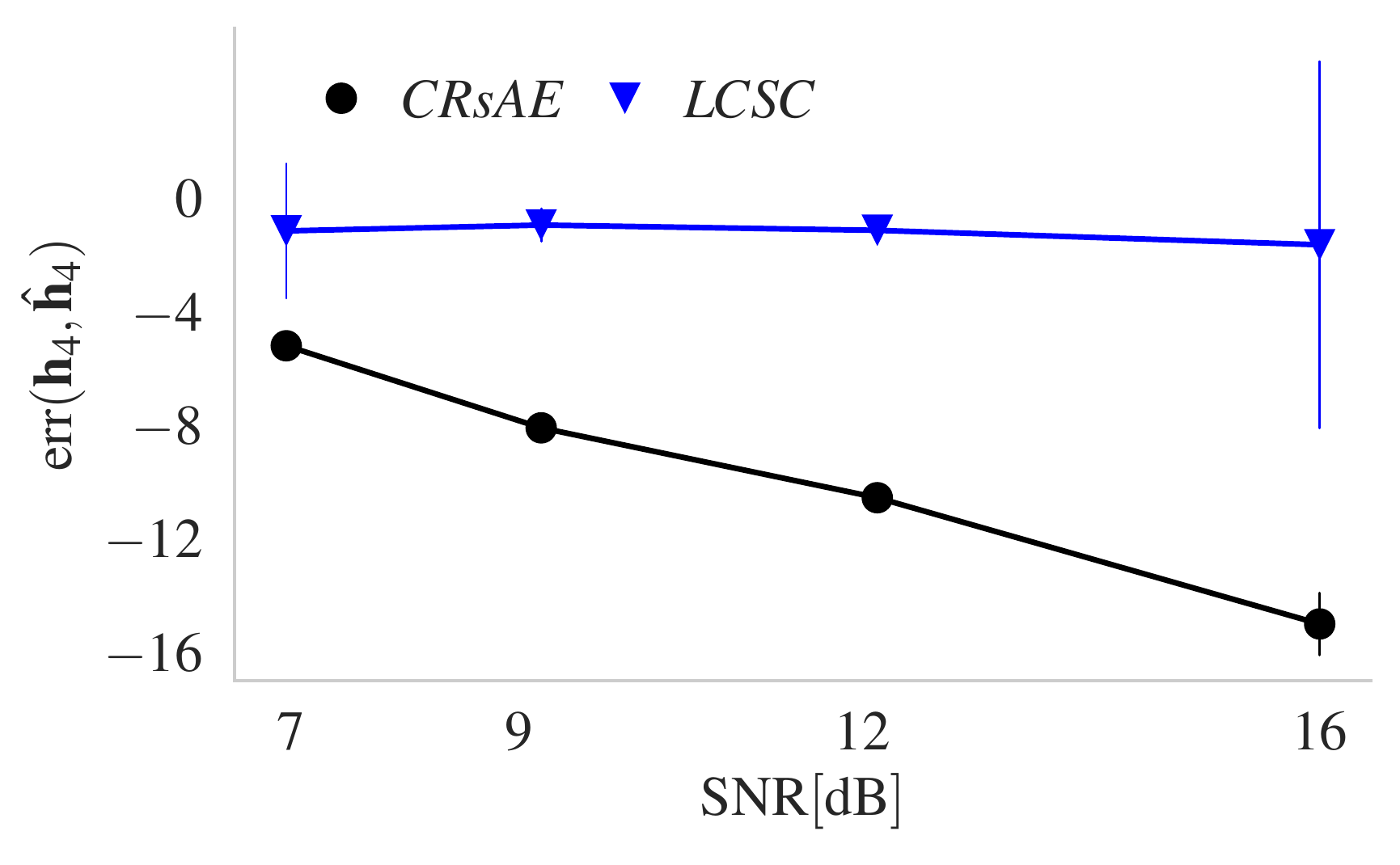}}
\end{minipage}
\vspace*{-8mm}
\caption{Plot of $\text{err}(\smlh_c,\hat{\smlh}_c)$ as a function of SNR. Compared to LCSC, CRsAE is able to recover the true filters and performs better as SNR increases.  Error $\text{err}(\smlh_c,\hat{\smlh}_c)$ is shown for only one of the filters as the curves for all the others were similar. The initial error is between $-3$ to$ -4$. This result is the average of $20$ independent trials. The vertical bars represent the variance.}
\label{fig:SNR}
\vspace*{-5mm}
\end{figure}

\vspace*{1mm}
\noindent \underline{\textbf{SNR analysis}}: We simulated data at four different equally-spaced SNRs in a log-scale ranging from $7$ to $16$ dB. We trained CRsAE and computed the measure from Eq.~(\ref{eq:distance_error}), which quantifies its ability to recover the filters used in the simulation, as a function of SNR. For training, we initialized the filters $\{\smlh^{\text{init}}_{c}\}_{c=1}^C$ by adding random Gaussian noise to the true filters $\{\smlh_c\}_{c=1}^C$ such that $\text{err}(\smlh_c,{\smlh}^{\text{init}}_c)$ was between $-3$ to $-4$ for all the filters. Figure~\ref{fig:SNR} depicts $\text{err}(\smlh_c,\hat{\smlh}_c)$ as a function of SNR. The error was averaged over $20$ independent experiments and is only shown for $c=4$ as the curves for all the other filters were similar. The vertical line presents the variance of the error, where we can see that the variance of CRsAE error is very small. The figure demonstrates that, compared to LCSC, CRsAE is able to recover the underlying filters and does so in a manner that is robust to noise. That is, CRsAE performs well for various level of the SNR and, as SNR increases, CRsAE learns filters that are closer to the true underlying ones. Interestingly, in spite of the fact that LCSC was able to reconstruct the noisy data very well, it failed to learn the underlying filters. Moreover, the filters that LCSC learned rarely correlated with the true filters as measured by Eq.~(\ref{eq:distance_error}). CRsAE, on the other hand, is able to perform denoising \emph{and} learn the true filters. We attribute the success of CRsAE to two factors. First, it is well-known in the dictionary-learning literature~\cite{Agarwal2016LearningSU} that, given a noisy initial dictionary, the success of dictionary learning depends on the ability of the sparse coding step to produce sparse codes that are close to those that generated the examples. In CRsAE,  we unfold for $T=180$ iterations, compared to LCSC that only ran on the order of $< 10$ iterations of ISTA. That's why the CRsAE encoder yielded sparse codes close to the true ones, while LCSC did not. Second, the fact that the encoder and decoder weights are tied in CRsAE means that it has fewer parameters to learn than LCSC with the same amount of data.

\vspace*{-4mm}

\begin{figure}[htb]
\begin{minipage}[b]{1.0\linewidth}
  \centering
  \centerline{\includegraphics[width=8.0cm]{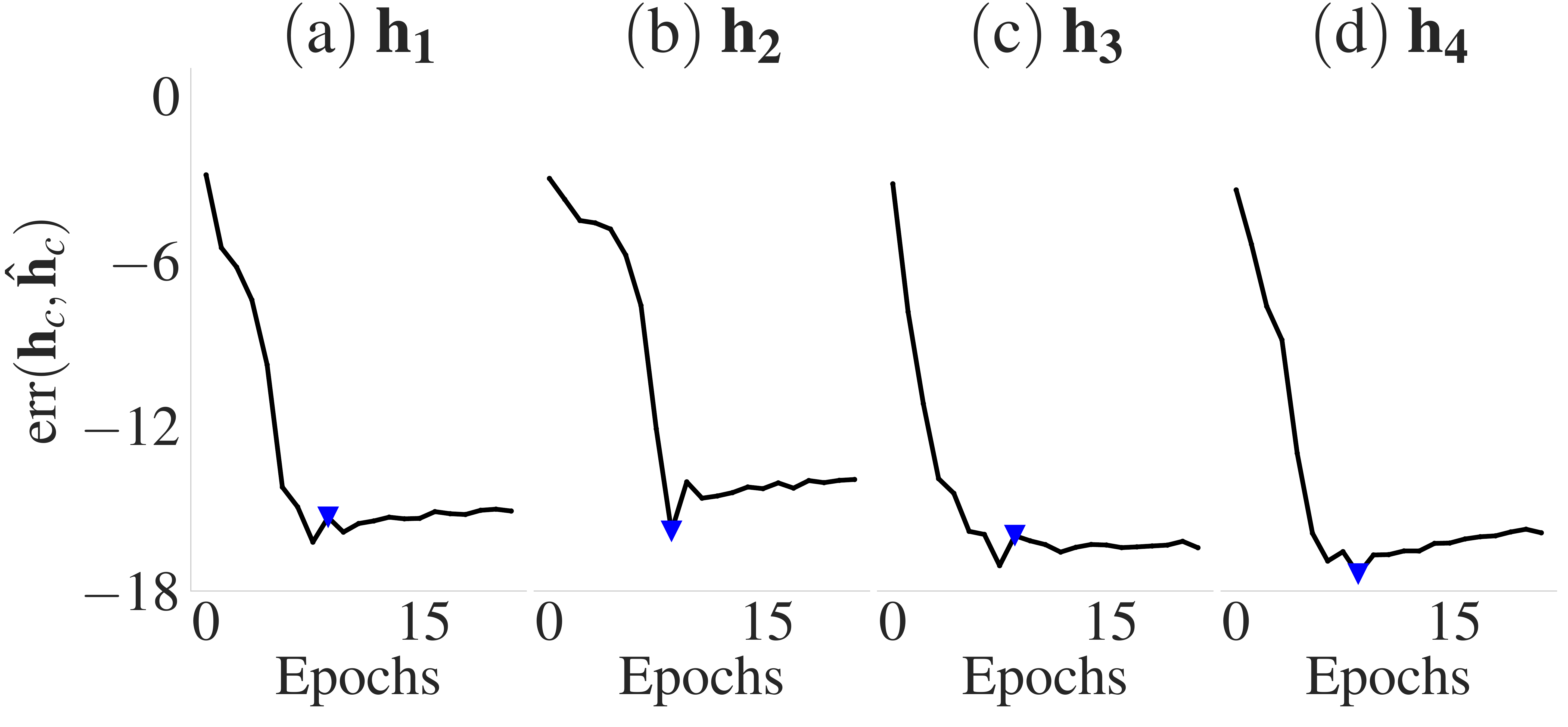}}
\end{minipage}
\vspace*{-7mm}
\caption{Error $\text{err}(\smlh_c,\hat{\smlh}_c)$ as a function of the number of epochs for the simulated dataset with $16$ dB SNR. The error $\text{err}(\smlh_c,\hat{\smlh}_c)$ reaches a value below $-14$ after only $6$ epochs, demonstrating that CRsAE converges fast. $\bigtriangledown$ denotes the epoch at which the validation loss is minimized.}
\label{fig:H_sim_err}
\vspace*{-3mm}
\end{figure}

\begin{figure}[htb]
\begin{minipage}[b]{1.0\linewidth}
  \centering
  \centerline{\includegraphics[width=8.0cm]{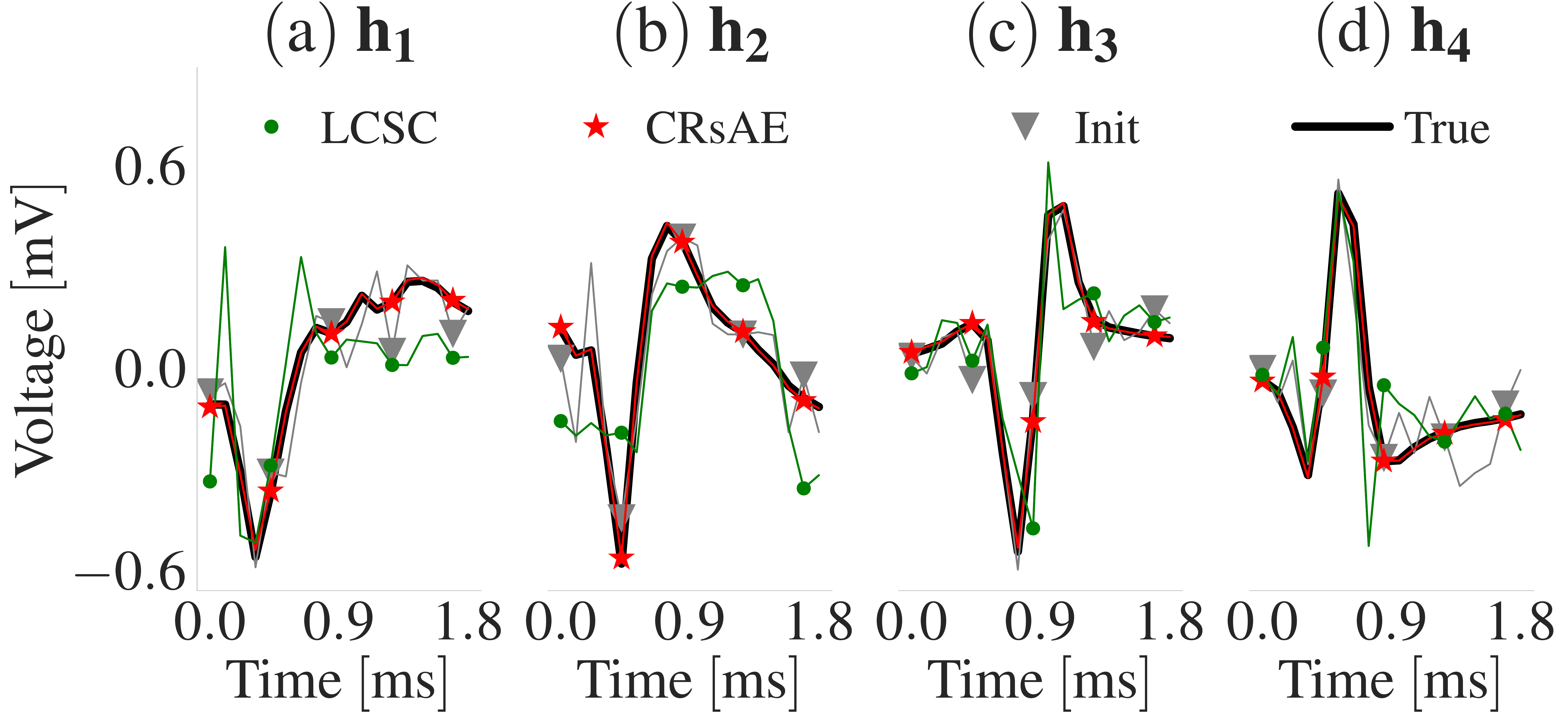}}
\end{minipage}
\vspace*{-7mm}
\caption{Filters learned by CRsAE using simulated data with a $16$ dB SNR. This highlights the ability of CRsAE to learn the true underlying filters (black). The gray curve ($\bigtriangledown$), red curve ($\star$), and green curve ($\circ$) correspond respectively to the initial filters, filters estimated by CRsAE, and the filters learned by LCSC.}
\label{fig:H_sim_learned}
\vspace*{-4mm}
\end{figure}

\vspace*{1mm}
\noindent \underline{\textbf{Dictionary learning}}: Figure~\ref{fig:H_sim_err} shows a plot of the distance error from Eq.~(\ref{eq:distance_error}) as a function of the number of epochs for one of the experiments from Figure~\ref{fig:SNR} with $16$ dB SNR. After $3$ to $4$ epochs, the distance between all learned and true filters decreases to less than $-14$. The learned dictionary corresponds to epoch 8, where the validation loss is minimized. The fast convergence of CRsAE is likely due to a) the low number of parameters to learn due to the sharing of parameters between the encoder and the decoder, and b) a large number of training examples. In the figure, the fact that the error increases slightly after it reaches a minimum is likely the effect of noise due to the stochastic nature of mini-batch gradient descent.

Figure~\ref{fig:H_sim_learned} shows the filters corresponding to the error plots from Figure~\ref{fig:H_sim_err}. In Figure~\ref{fig:H_sim_learned}, gray color (denoted by $\bigtriangledown$) indicates the initial filters. The true filters are shown in black. CRsAE is able to learn filters (red color denoted by $\star$) that are indistinguishable from the true filters, but LCSC (denoted by $\circ$  and depicted in green) fails the task. Together, Figures~\ref{fig:SNR},~\ref{fig:H_sim_err} and~\ref{fig:H_sim_learned} demonstrate the ability of CRsAE to perform convolutional dictionary learning, and its robustness to noise.

\begin{figure}[htb]
\begin{minipage}[b]{1.0\linewidth}
  \centering
  \centerline{\includegraphics[width=8.0cm]{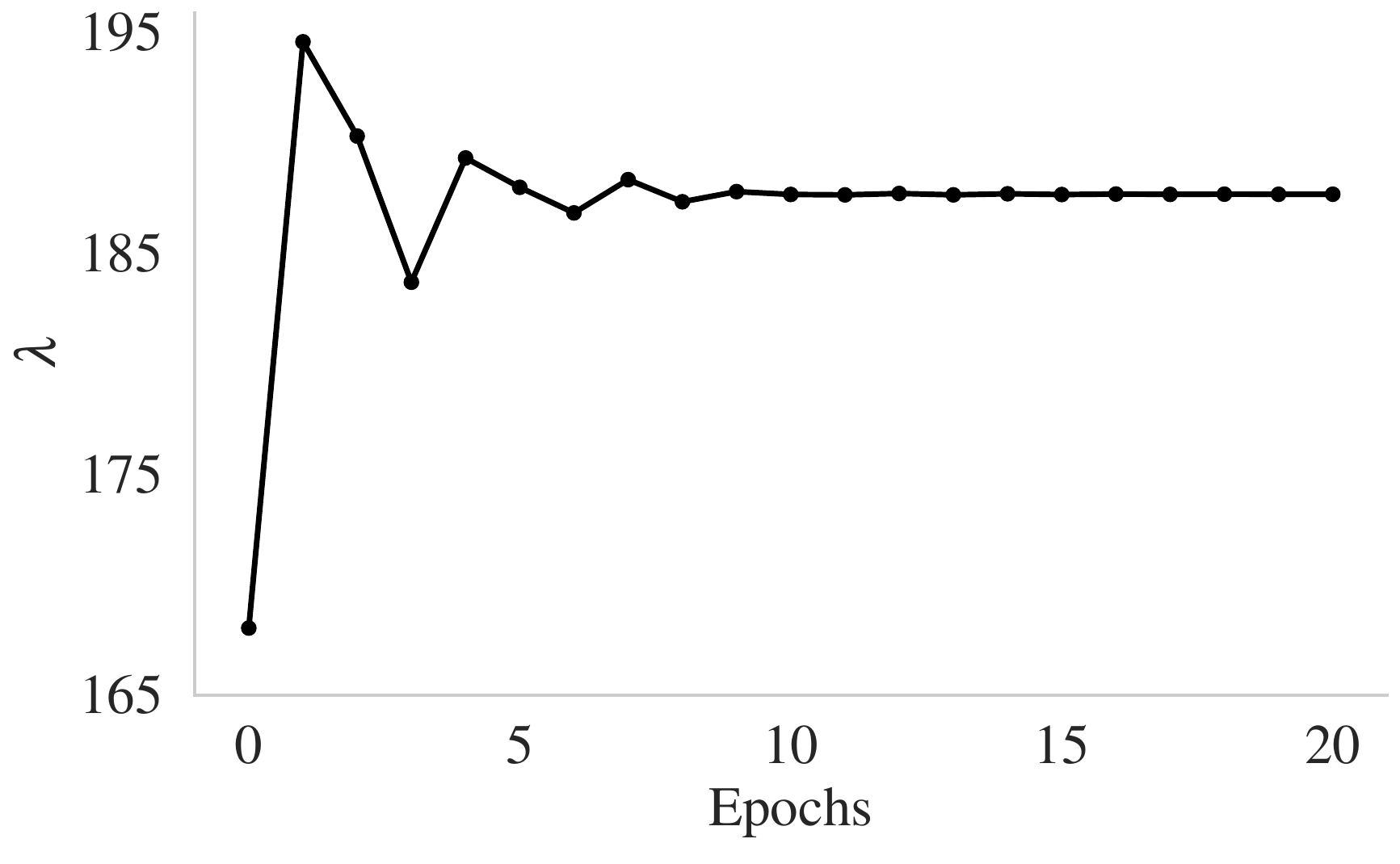}}
\end{minipage}
\vspace*{-8mm}
\caption{The regularization parameter $\lambda$ learned by CRsAE in each epoch of training for a trial for $16$ dB SNR.}
\label{fig:lambda}
\end{figure}

\vspace*{-1mm}
\noindent \underline{\textbf{Learning the regularization parameter} (ReLU bias)}: The EM-inspired CRsAE architecture lets us automatically train and learn the regularization parameter using noisy data. Figure~\ref{fig:lambda} shows the values of $\lambda$ for CRsAE as a function of epochs in a simulation with SNR of $16$ dB. The figure shows that $\lambda$ for CRsAE converges after $4$ to $5$ epochs. We initialized $\lambda$ as detailed in Section~\ref{sec:training} to a value of $168.04$. The learned $\lambda$ converged to $187.68$. For LCSC, $\lambda$ for all filters converged to a very small value, resulting in over-fitting the noise in the dataset. We note that for a fair comparison, we trained LCSC using the same reconstruction loss as CRsAE. However, even if we had optimized the LCSC loss function, $\lambda$ would have behaved the same due to the absence of the regularization on the parameter as in the Bayesian loss function from Eq.~(\ref{eq:loss_lambda}), specifically the second term. 

\vspace*{1mm}
\noindent \underline{\textbf{Speed analysis}}: We compared the speed of CRsAE applied to CDL with Sporco. For fairness of comparison, we compared the speeds given a fixed desired dictionary learning accuracy. 

\begin{figure}[htb]
\begin{minipage}[b]{1.0\linewidth}
  \centering
  \centerline{\includegraphics[width=8.0cm]{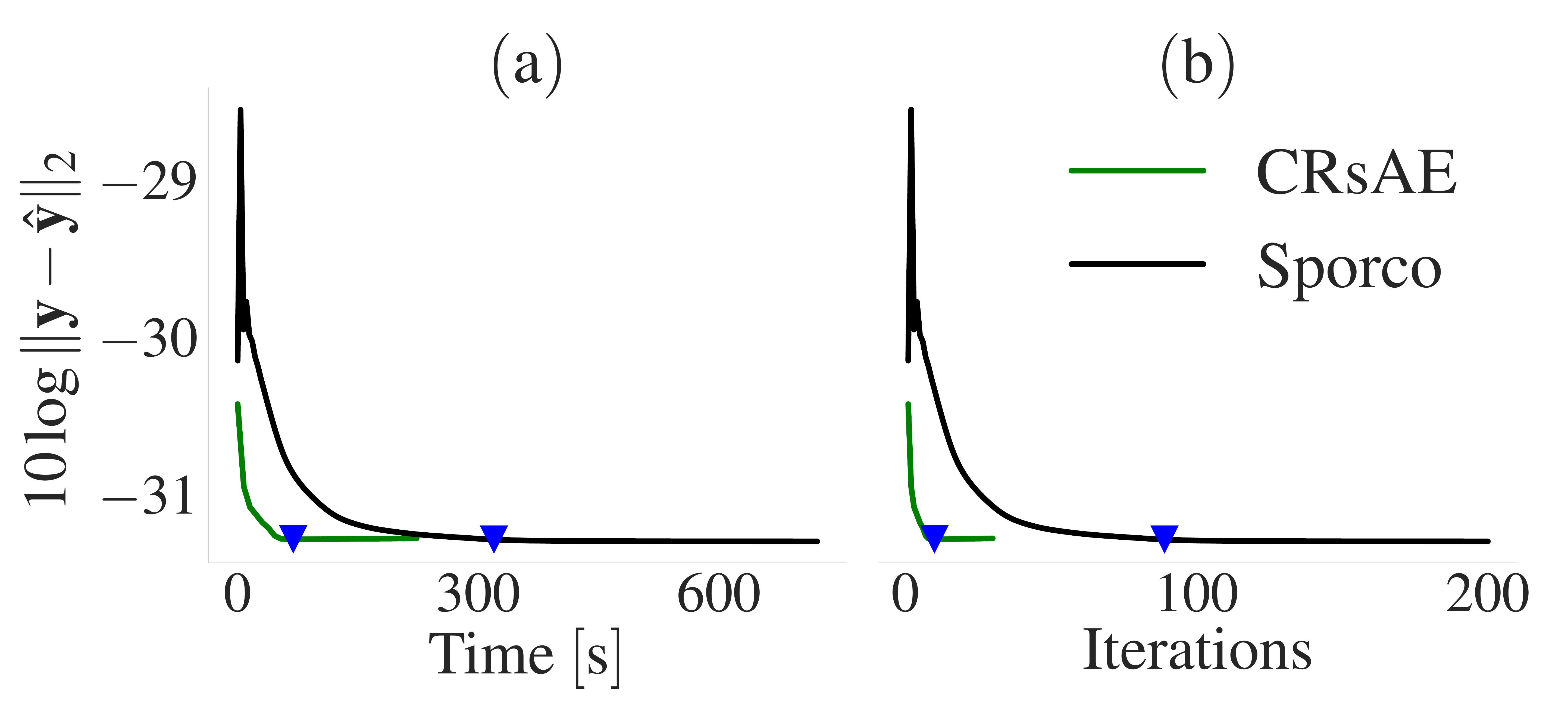}}
\end{minipage}
\vspace*{-7mm}
\caption{Comparison of CRsAE to Sporco in terms of dictionary learning speed on simulated data with $16$ dB SNR. In this example, CRsAE performs dictionary learning $5$x faster than Sporco. $\bigtriangledown$ denotes the iteration/time at which the methods attain the minimum CRsAE validation loss.}
\label{fig:speed_analysis}
\vspace*{-4mm}
\end{figure}

We used simulated data at $16$ dB SNR and set $\lambda$ to the learned value from Section~\ref{sec:sim}. We used the same regularization parameter ($\lambda \sigma^2 = 187.68\ (0.024)^2 \approx 0.11$ for SNR of $16$ dB) for Sporco. Other hyperparameters of Sporco $(\rho, L)$ were tuned to $(4,10^5)$ by random grid-search guided by the hyperparameters used in~\cite{garcia-2018-convolutional}. We ran CRsAE and Sporco for 30 and 200 iterations, respectively. Here, $1$ iteration refers to going over all of the dataset, i.e., an epoch in neural-networks terminology. Figure~\ref{fig:speed_analysis}(a) plots the signal reconstruction error for CRsAE and Sporco as a function of time. The figure shows that CRsAE performs CDL $5$x faster than Sporco. Indeed, CRsAE took $69.27$ s to train, and Sporco reached the same accuracy as CRsAE in $319.52$ s. Figure~\ref{fig:speed_analysis}(b) shows that CRsAE took $7$ epochs to train, and the training loss reached its minimum in $10$ epochs (denoted by $\bigtriangledown$). Sporco took $89$ iterations to reach the same accuracy/loss as CRsAE. The first row of Table~\ref{tab:speed} summarizes the speed comparison between CRsAE and Sporco. We chose to plot the signal reconstruction error, as opposed to Eq.~(\ref{eq:distance_error}), to demonstrate that CRsAE can also perform the signal reconstruction. Both CRsAE and Sporco converged to the filters in Figure~\ref{fig:H_sim_learned}.

%%%%%%%%%%%%%%%%%%%%%%%%%%%%%%%%%%
\begin{table} 
\caption{Details of Speed Analysis}
\label{tab:speed}
\vspace*{-2mm}
  \centering
  \begin{tabular}{llllll}
        &   & CRsAE & Sporco & CBP\\ \midrule
    \multirow{2}{*}{Dictionary Learning}  & \multicolumn{1}{l}{runtime} & \multicolumn{1}{l}{$\textbf{69.27}$ \textbf{s}} & \multicolumn{1}{l}{$319.52$ s} &  \\ \cline{2-4}
                                 & \multicolumn{1}{l}{iterations} & \multicolumn{1}{l}{$\textbf{10}$} & \multicolumn{1}{l}{$89$} &   \\ \midrule
Spike Sorting & runtime & $\textbf{0.93}$ \textbf{s} &   & $17$ hours \\ \bottomrule
   \\
  \end{tabular}
\vspace*{-7mm}
\end{table}

%%%%%%%%%%%%%%%%%%%%%%%%%%%%%%%%%%
\vspace*{-4mm}
%%%%%%%%%%%%%%%%%%%%%%%%%%%%%%%%%%
%\vspace*{-3mm}
\subsection{Spike Sorting: Real Data}\label{sec:real}

\noindent \underline{\textbf{Extracellular data from rat Hippocampus}}: The dataset consists of extracellular tetrode recordings from the rat Hippocampus with simultaneous intracellular recording. The intracellular recording provides the ground-truth data that is used to assess the ability of CRsAE to identify spikes successfully. The reader can find additional details about these data in~\cite{HarrisKd2000Aots}. The goal is to sort the spikes that appear in the extracellular recordings. Spike sorting is an important source-separation problem in computational neuroscience~\cite{lewicki1998review}. It refers to the process of identifying the time of occurrence of action potentials (filters) in an extracellular voltage recording and their assignment to a neuron (source). Assuming the extracellular data follows the model of Eq.~(\ref{eq:DTconv}), we cast spike sorting as a CDL problem and apply CRsAE to it.

\noindent \underline{\textbf{Pre-processing}}: We high-passed filter the raw data (Channel 0) at $400$ Hz and whiten the noise. To minimize boundary effects, we picked a large window length of $6$ s, resulting in a total of $24$ windows (examples). We set the length of the filters to be learned to $3.5$ ms. The sampling rate of the extracellular data was $f_s$$=$$10$ kHz. Therefore, each window had length $N$$=$$60{,}000$ samples, and each filter had length $K$$=$$35$ samples. In the recording, there were mainly two neurons spiking~\cite{HarrisKd2000Aots}, so we set $C$$=$$2$. The data were divided by the maximum absolute value of the recording. Prior to normalization, we estimated the standard deviation of the noise from ``silent'' periods in the recording, yielding a value of $1$ in fractions of mV. We do not specify the units more precisely because the data are made available to us in an already-normalized form. Following normalization by the maximum absolute value of the signal, we tuned $\sigma$ to a value of $0.03$. For training, we split the data into $21$ examples for training and $3$ examples for validation. To improve the learning performance, the training set was tripled by data augmentation (see Section~\ref{sec:training}), resulting in $63$ examples. The second column of Table~\ref{tab:dataset1d} summarizes this dataset and the parameters used for training. The depth of the network is $T+1 = 601$ layers, and the number of channels at every layer is $C=2$, the number of neurons. The width of each channel is $N-K+1 = 59{,}966$.

\begin{figure}[htb]
\begin{minipage}[b]{1.0\linewidth}
  \centering
  \centerline{\includegraphics[width=8.0cm]{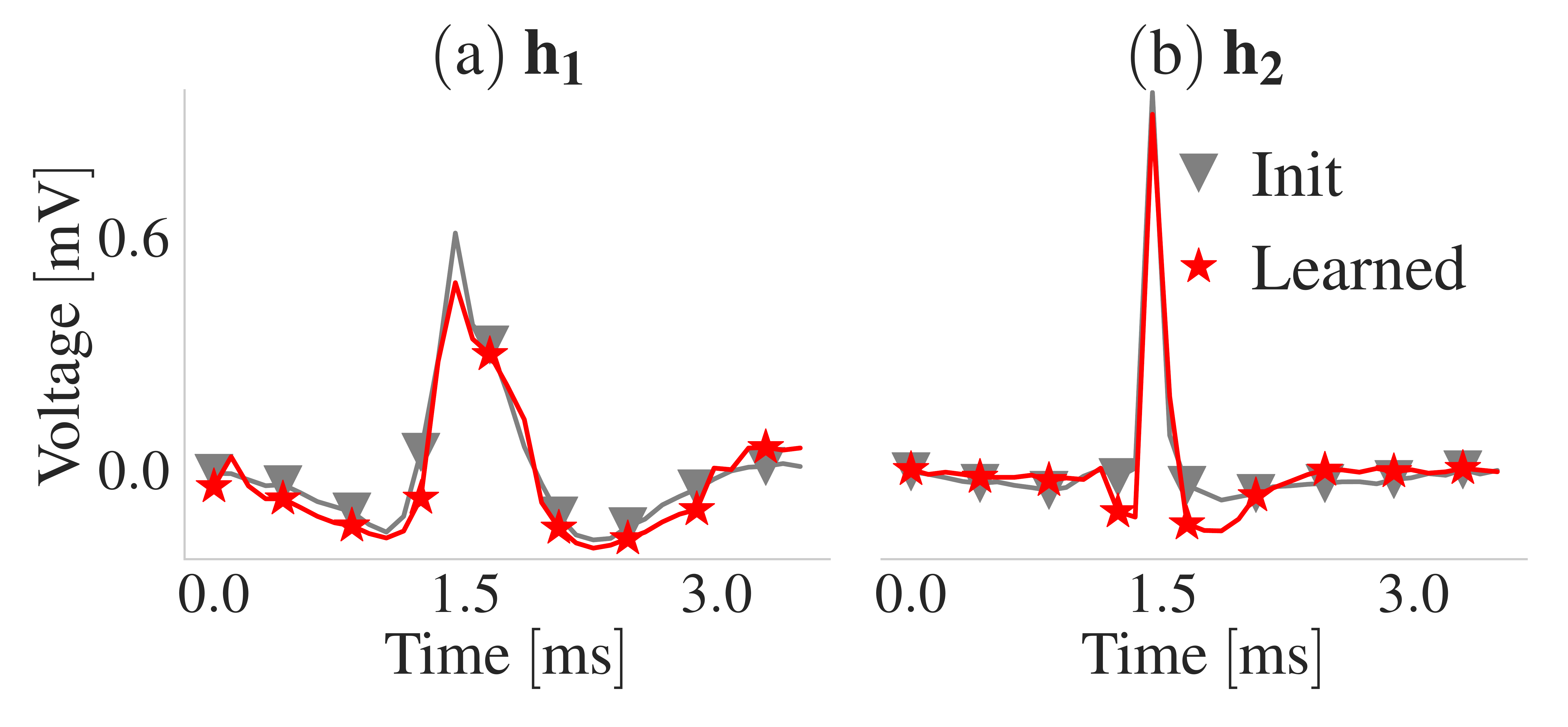}}
\end{minipage}
\vspace*{-8mm}
\caption{Filters estimated by CRsAE applied to the extracellular tetrode recordings. The gray ($\triangledown$) curve corresponds to the initial filters. The filters learned by CRsAE are in red ($\star$). The figure shows that CRsAE is able to learn filters that resemble the action potentials of a neuron.}
\label{fig:H_real}
\end{figure}

\noindent \underline{\textbf{Dictionary learning}}: To initialize the filters, we used the following standard procedure~\cite{ekanadham2014unified}. We spotted the location of potential spikes using a pre-computed threshold and windowed the data around the spotted positions. Then, we computed the two first principal components of the matrix whose rows comprise the obtained windows. Finally, we performed k-means clustering on the windows in principal-component space and picked the initial filters as the center of the clusters. We initialized $\lambda$ to $161.21$, and it converged to $219.25$ upon training. Figure~\ref{fig:H_real} shows the filters before and after training CRsAE. The initial and learned filters are in gray (denoted by $\triangledown$) and in red (denoted by $\star$), respectively. The figure shows that CRsAE is able to learn filters that resemble the action potentials of a neuron. These filters are interpretable, in the sense that they are those that allow us to best represent the data in the form of the convolutional generative model of Eq.~(\ref{eq:DTconv}).

\noindent \underline{\textbf{Spike sorting}}: In this experiment, we did not have access to the ground-truth filters (true action potentials) from the extracellular recordings. However, the intracellular voltage recording was available for the neuron corresponding to the filter $\smlh_1$. We used the intracellular data to assess how well CRsAE can perform spike sorting. Unlike traditional methods for spike sorting that use principal component analysis to identify distinct templates within the data, CRsAE uses raw extracellular data. We compared CRsAE to CBP. Unlike CRsAE, CBP does not perform CDL: it simply performs CSC to identify the time of occurrence of filters \emph{given} the filters.

We quantify the ability of an algorithm to perform spike sorting by comparing the spike trains it estimates to the true spikes (sparse codes). The true spikes are provided through intracellular voltage recordings. Similar to \cite{ekanadham2014unified}, for a given threshold, we compute the proportion of true (intracellular) spikes that are not identified correctly by CRsAE (true miss), as well as the proportion of spikes identified by CRsAE for which there is no true matching spike (false alarm). In practice, to identify a spike, given the neuron of interest $c$, we reconstruct the component of the data corresponding only to the neuron of interest. Then, we use a threshold and identify any non-zero spiking activity greater than the threshold to be a spike. A low threshold value results in identifying background noise or other artifacts as spikes (false alarm), whereas a high threshold may result in missing spikes (true miss).

\begin{figure}[htb]
\begin{minipage}[b]{1.0\linewidth}
  \centering
  \centerline{\includegraphics[width=8.0cm]{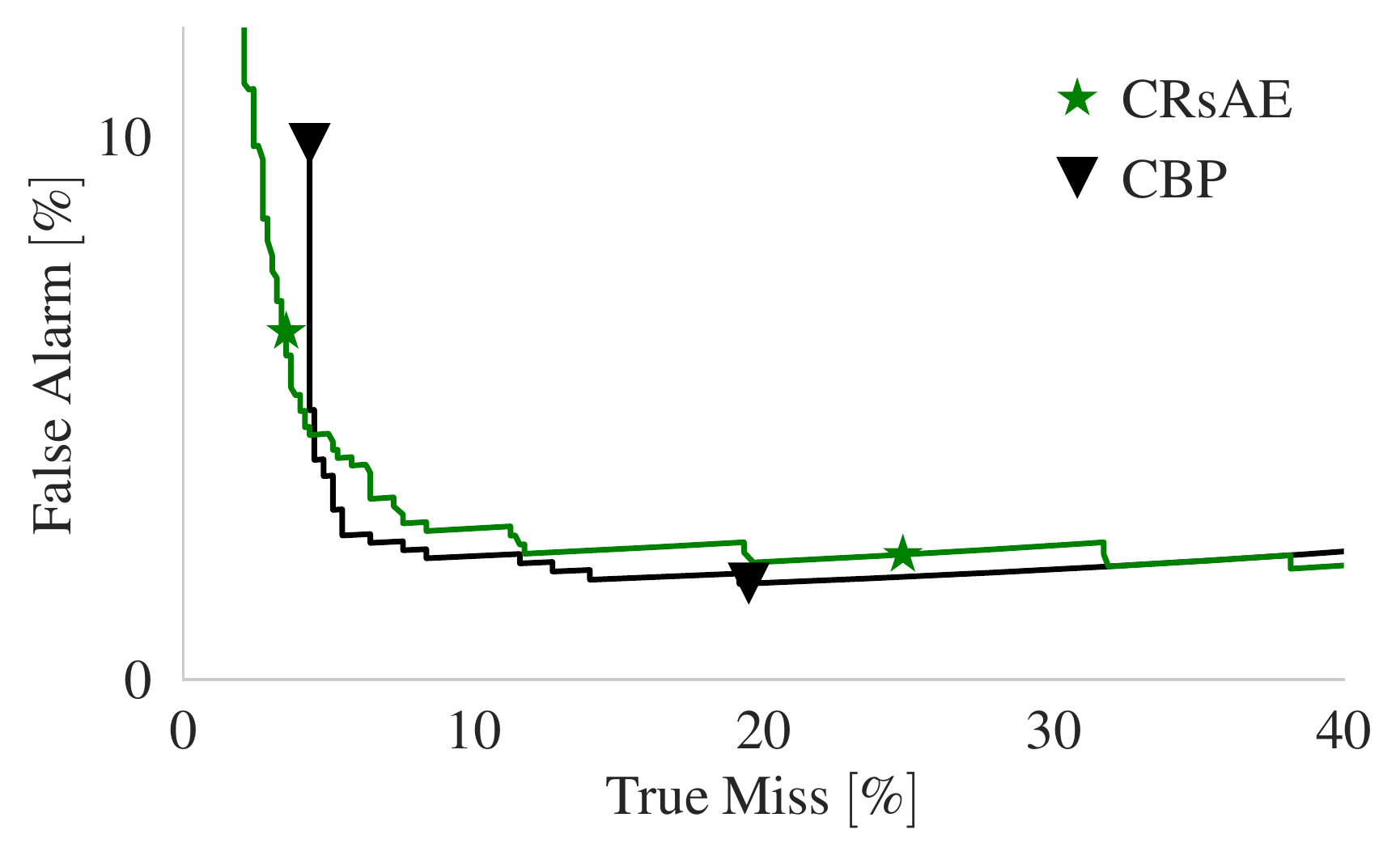}}
\end{minipage}
\vspace*{-8mm}
\caption{Comparison between CRsAE and CBP applied to spike sorting. The trade-off curves, between \textit{true missed} and \textit{false alarm} proportions, obtained by using CRsAE (green $\star$) and CBP (black $\triangledown$) to sort the spikes from the neuron for which intracellular data are available. The figure demonstrates that CRsAE is competitive with CBP.}
\label{fig:miss_false}
\end{figure}

Figure~\ref{fig:miss_false} shows the trade-off between the \textit{false alarm} proportion and the \textit{true miss} proportion as a function of threshold and highlights the competitive performance of CRsAE compared to CBP. We also compared the speed of CRsAE to that of CBP for spike sorting of the simulated data lasting $17$ minutes. CRsAE learned the filters in $69.27$ s and performed spike sorting in only $0.93$ s. The implementation of CBP at our disposal is not efficient enough to be applied to a $17$-minute-long recording. We estimated that CBP would take $17$ hours to perform spike sorting given the filters. We estimated this timing by running CBP on $1$, $2$, $4$, and $8$ s of the data and observed a linear increase in the timing CBP needed for spike sorting~\cite{ekanadham2014unified}. The second row of Table~\ref{tab:speed} compares the speed of the two algorithms of the two algorithms applied to spike sorting.

% spike sorting results filename: paper_real_harris_filter2_fdim30_lam_lowersigma

\vspace*{-4mm}
\vspace*{-8mm}

\begin{table} 
\caption{Details of Datasets and Training Parameters for Image Experiments.}
\label{tab:dataset2d}
\vspace*{-6mm}
  \centering
  \setlength\tabcolsep{5pt}
  \begin{tabular}{cccc}
           &  \\  \midrule
     \multirow{2}{2cm}{\# filters $C$} & tied &  \multicolumn{2}{c}{untied} \\  
     & $64$ & \multicolumn{2}{c}{$3$$\times$$64$$=$$192$}\\ \midrule
    Filter size $K$ & \multicolumn{3}{c}{$7$$\times$$7$}  \\  \midrule
    Strides & \multicolumn{3}{c}{$5$} \\  \midrule
    Patch size & \multicolumn{3}{c}{$128$$\times$$128$} \\  \midrule
    \# training examples & \multicolumn{3}{c}{$5{,}717$ VOC}\\ \midrule
    \# testing examples  & \multicolumn{3}{c}{$9$}\\ \midrule
    \multirow{2}{2.7cm}{\# trainable parameters} & EM or LS & hyp & EM or LS-untied\\
     & $3{,}200$ & $3{,}136$ & $9{,}472$\\  \midrule
     Encoder layers $T$ & \multicolumn{3}{c}{$30$}\\  \midrule
     Batch size $B$ & \multicolumn{3}{c}{$1$} \\ \midrule
     $L$ & \multicolumn{3}{c}{$10$} \\ \midrule
    \multirow{2}{0.5cm}{$\lambda_{\text{init}}$} & EM & hyp & LS\\
     & 20 &  0.15 & 0.015 \\ \midrule
    ${\smlh}^{\text{init}}$ & \multicolumn{3}{c}{Random Gaussian} \\ \midrule
  $\eta_{\smlh}$ & \multicolumn{3}{c}{$0.01$}\\  \midrule
    \multirow{2}{0.3cm}{$\eta_{\lambda}$} & EM & hyp & LS\\
     & $0.1$ & - & $0.01$\\  \midrule
  $\eta_{\smlh}$ decay & \multicolumn{3}{c}{$0.7$} \\  \midrule
  $\eta_{\smlh}$ step & \multicolumn{3}{c}{$10$ epochs}\\  \midrule
  \multirow{2}{0.3cm}{$\eta_{\lambda}$ decay} & EM & hyp & LS\\
  & - & - & $0.7$\\  \midrule
  \multirow{2}{0.3cm}{$\eta_{\lambda}$ step} & EM & hyp & LS\\
  & - & - & $10$ epochs\\ \bottomrule
  \end{tabular}
\vspace*{-2mm}
\end{table}

\begin{figure*}[h]
	\vspace*{-4mm}
	%%%%%% CRsAE %%%%%%
	\begin{minipage}[b]{0.98\linewidth}
		\centering
		\tikzstyle{input} = [coordinate]
		\tikzstyle{output} = [coordinate]
		\tikzstyle{pinstyle} = [pin edge={to-,thin,black}]
		\begin{tikzpicture}[auto, node distance=2cm,>=latex']
		cloud/.style={
			draw=red,
			thick,
			ellipse,
			fill=none,
			minimum height=1em}
		% We start by placing the blocks
		\node [input, name=input] {};
		
		\node [rectangle, fill=none, node distance=0.001cm, right of=input] (A) {$\includegraphics[width=0.25\linewidth]{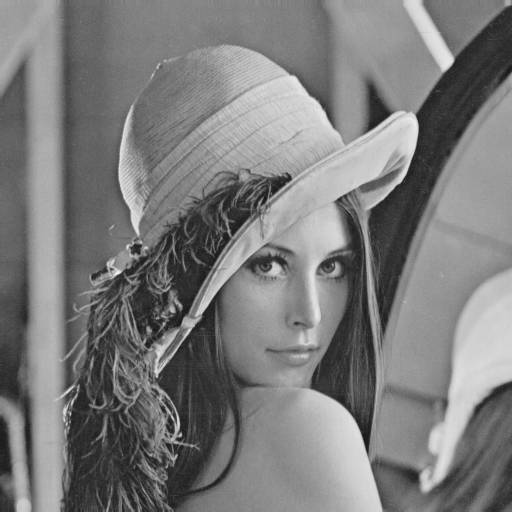}$};
		\node [rectangle, fill=none, node distance=4.5cm, right of=A] (B) {$\includegraphics[width=0.25\linewidth]{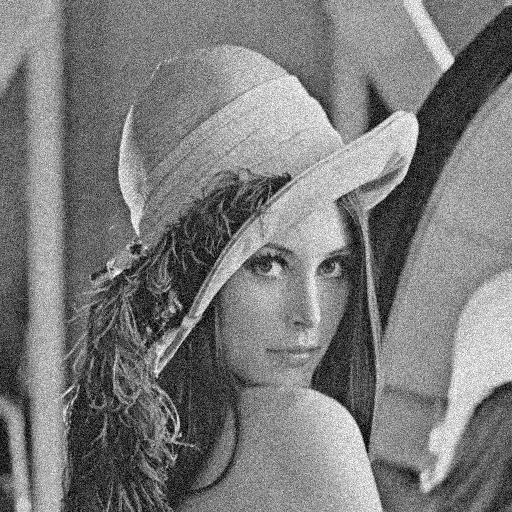}$};
		\node [rectangle, fill=none, node distance=4.5cm, right of=B] (C) {$\includegraphics[width=0.25\linewidth]{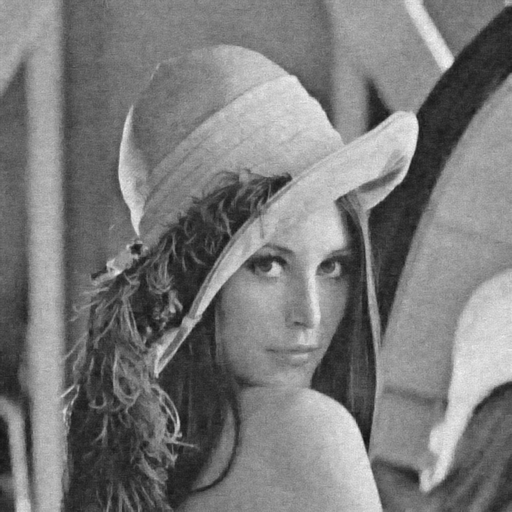}$};
		\node [rectangle, fill=none, node distance=4.5cm, right of=C] (D) {$\includegraphics[width=0.25\linewidth]{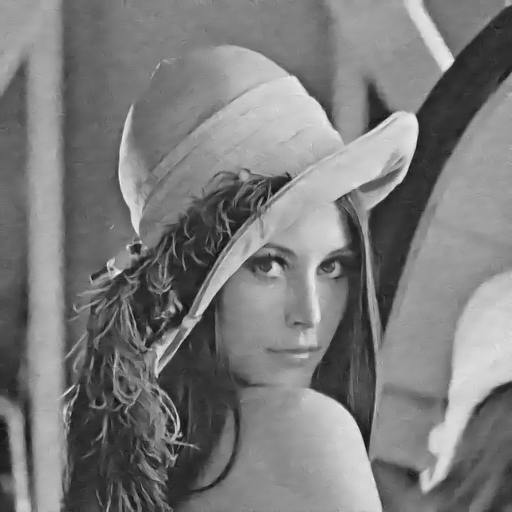}$};

		\node [rectangle, fill=none,  node distance=2.45cm,  above of=A] (text) {(a) Original};
		\node [rectangle, fill=none,  node distance=2.45cm,  above of=B] (text) {(b) Noisy};
		\node [rectangle, fill=none,  node distance=2.45cm,  above of=C] (text) {(c) CRsAE-$\lambda_{\text{EM}}$-shallow};
		\node [rectangle, fill=none,  node distance=2.45cm,  above of=D] (text) {(d) CRsAE-$\lambda_{\text{EM}}$-untied-msssim};	
		
		\end{tikzpicture}
	\end{minipage}
	\vspace*{-2mm}
	\caption{Visualizing Lena for denoising performance. (a) Original, (b) noisy, images denoised by (c) CRsAE-$\lambda_{\text{EM}}$-shallow, and (d) CRsAE-$\lambda_{\text{EM}}$-untied-msssim.}
	\label{fig:lena}
	\vspace*{-5mm}
\end{figure*}
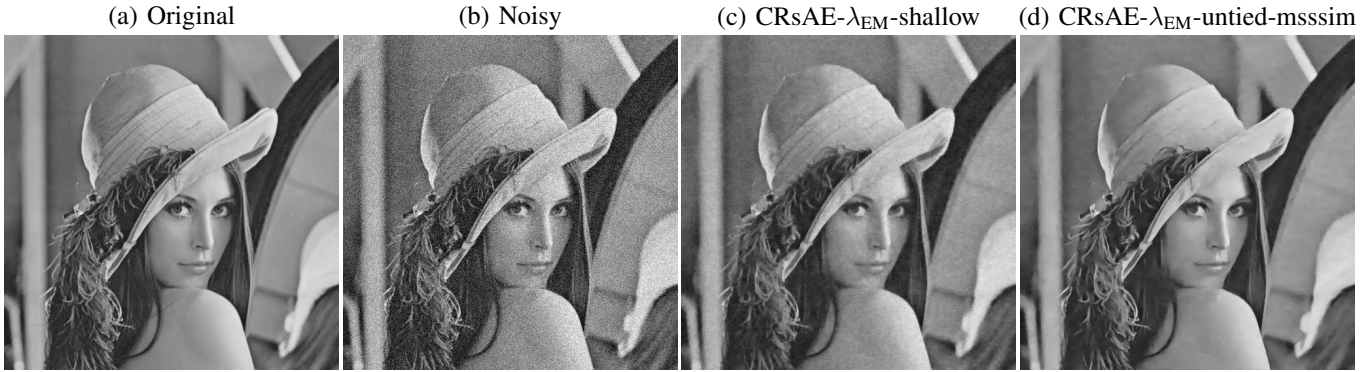

\begin{table*}[htb]
	\renewcommand{\arraystretch}{1.3}
	\caption{Denoising Performance of CRsAE on test images quantified by PSNR. The number of trainable parameters for each network is shown in the last column. LCSC results are from~\cite{SreterHillel2018LCSC}.}
	\label{tab:psnr}
	\centering
	\begin{tabular}{c|c|c|c|c|c|c|c|c|c||c}
	\hline
	\hline
		& Lena & Barbara & Boat & Couple & Fgpr & Hill & House & Man & Peppers & \# Parameters\\
		\hline
CRsAE-$\lambda_{\text{EM}}$-shallow & 29.74 & 26.19 & 28.89 & 28.69 & 26.69 &  28.50 &  29.76 & 28.51 & 
28.39 & 3{,}200\\
CRsAE-$\lambda_{\text{hyp}}$ & 30.17 & 27.57 & 29.21 & 28.81 & 27.08 &  28.35 &  30.36 & 28.56 & 29.41 & 3{,}136\\
CRsAE-$\lambda_{\text{LS}}$ & 30.22 & 27.65 & 29.39 & 29.17 & 27.23 &  28.84 &  30.23 & 28.98 & 29.51 & 3{,}200\\
CRsAE-$\lambda_{\text{EM}}$ & 30.56 & 28.10 & 29.61 & 29.35 & 27.42 &  29.04 &  30.58 & 29.21 & 29.83 & 3{,}200\\
CRsAE-$\lambda_{\text{LS}}$-untied & 31.44 & 28.72 & 30.17 & 29.95 & 27.46 &  29.45 &  31.63 & 29.67 & 30.67 &  9{,}472\\
CRsAE-$\lambda_{\text{EM}}$-untied & 31.83 & 28.92 & 30.41 & 30.20 & {\bf 27.82} &  29.67 &  32.11 & 29.89 & 30.95 & 9{,}472\\
CRsAE-$\lambda_{\text{EM}}$-untied-msssim & \bf{32.11} & {\bf 28.93} & {\bf 30.46} & {\bf 30.26} & 27.73 &  29.66 & 32.48 & 29.92 & {\bf 31.14} &  9{,}472\\
		LCSC & {\bf 32.11} & 28.91 & 30.30 & 30.14 & 27.44 & {\bf 30.23} & {\bf 32.55} & {\bf 30.29} &  30.65 & 9{,}472\\ \bottomrule
	\end{tabular}
	\vspace*{-3mm}
\end{table*}

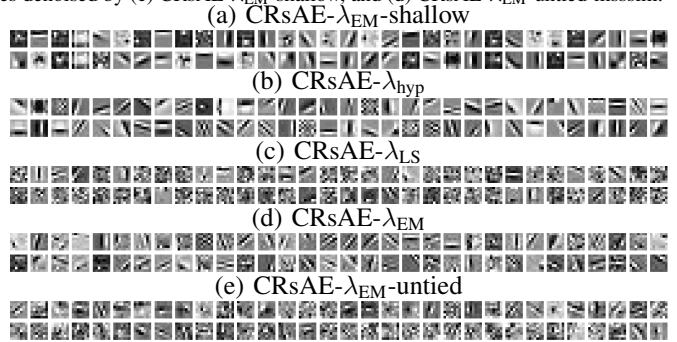
\begin{figure}[h]
	\vspace*{-4mm}
	%%%%%% CRsAE %%%%%%
	\begin{minipage}[b]{1.0\linewidth}
		\centering
		\tikzstyle{input} = [coordinate]
		\tikzstyle{output} = [coordinate]
		\tikzstyle{pinstyle} = [pin edge={to-,thin,black}]
		\begin{tikzpicture}[auto, node distance=2cm,>=latex']
		cloud/.style={
			draw=red,
			thick,
			ellipse,
			fill=none,
			minimum height=1em}
		% We start by placing the blocks
		\node [input, name=input] {};
		
		\node [rectangle, fill=none, node distance=0.001cm, right of=input] (A) {$\includegraphics[width=1\linewidth]{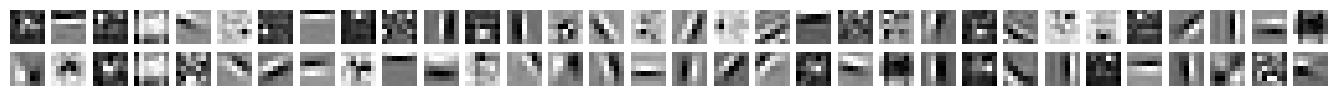}$};
		\node [rectangle, fill=none, node distance=0.9cm, below of=A] (B) {$\includegraphics[width=1\linewidth]{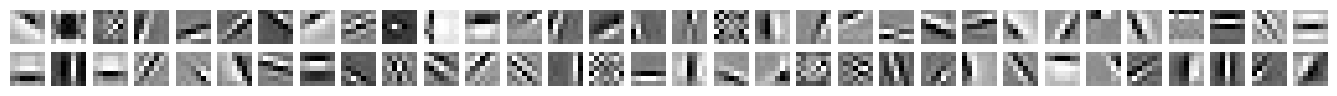}$};
		\node [rectangle, fill=none, node distance=0.9cm, below of=B] (C) {$\includegraphics[width=1\linewidth]{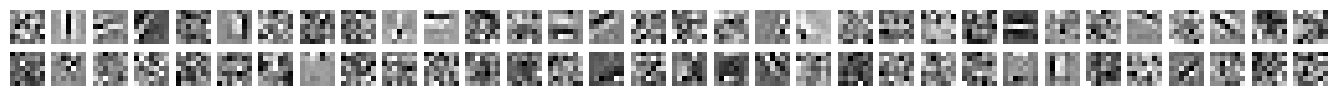}$};
		\node [rectangle, fill=none, node distance=0.9cm, below of=C] (D) {$\includegraphics[width=1\linewidth]{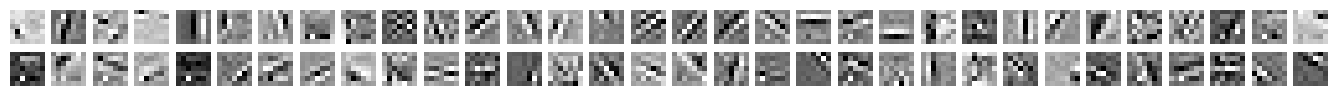}$};
		\node [rectangle, fill=none, node distance=0.9cm, below of=D] (E) {$\includegraphics[width=1\linewidth]{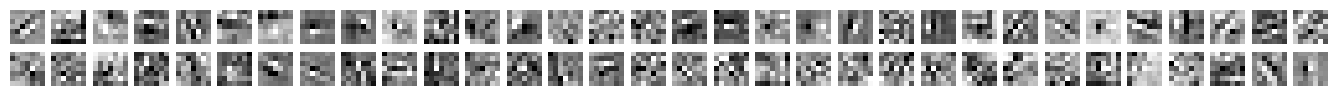}$};
		
		\node [rectangle, fill=none,  node distance=0.44cm,  above of=A] (text) {(a) CRsAE-$\lambda_{\text{EM}}$-shallow};
		\node [rectangle, fill=none,  node distance=0.44cm,  above of=B] (text) {(b) CRsAE-$\lambda_{\text{hyp}}$};
		\node [rectangle, fill=none,  node distance=0.44cm,  above of=C] (text) {(c) CRsAE-$\lambda_{\text{LS}}$};
		\node [rectangle, fill=none,  node distance=0.44cm,  above of=D] (text) {(d) CRsAE-$\lambda_{\text{EM}}$};     
		\node [rectangle, fill=none,  node distance=0.44cm,  above of=E] (text) {(e) CRsAE-$\lambda_{\text{EM}}$-untied};   
		\end{tikzpicture}
	\end{minipage}
	\vspace*{-8mm}
	\caption{Learned filters at the decoder.}
	\label{fig:filters}
	\vspace*{-6mm}
\end{figure}

%crsae_kernel8_conv64_stride7_crop64_lrp1_notTied_fista3_sigmap04

%%%%%%%%%%%%%%%%%%%%%%%%%%%%%%%%%%

\vspace*{-4mm}
\subsection{Image Denoising}\label{sec:image}

We trained CRsAE to denoise images from the PASCAL VOC image dataset~\cite{pascal-voc-2012}, consisting of $J=5{,}717$ training images. We used a set of $9$ commonly used test images for evaluation purposes, where the peak signal-to-noise-ratio (PSNR) is used as the evaluation criterion. We trained CRsAE in supervised settings. The supervised setting assumes access, during training, to pairs that consist of a noisy and the associated clean image.

We learned $C = 64$ filters, each of size $7$$\times$$7$. We used convolutions with a stride of $5$ pixels in each dimension. In order to capture the patterns appearing in shifts missed by the strides, we followed a similar approach to~\cite{rethinking} for augmentation and reconstruction. In total, we compared seven networks. We considered three networks to compare different methods of training the bias. CRsAE-$\lambda_{\text{EM}}$ learned $\lambda \in \R^{64}$ (one for each filter) through the EM-inspired approach. CRsAE-$\lambda_{\text{hyp}}$ treated $\lambda$ as a hyperparameter and tuned it by grid search (we used the estimate $\lambda = \frac{\sqrt{2\log( C\times N_e)}}{\sigma}$~\cite{Chen1998AtomicDB} and tuned $\sigma$ by grid search in a neighborhood of the noise level of the noisy dataset). CRsAE-$\lambda_{\text{LS}}$ learned $\lambda$ using the standard approach in deep learning, namely by minimizing the global reconstruction loss~\cite{SreterHillel2018LCSC,rethinking,mardani2018}. For the cases in which $\lambda$ was tuned, we absorbed $\sigma^2$ into $\lambda$ when reporting numbers. Similarly, for the case in which $\lambda$ was trained through a reconstruction loss, we absorbed $\sigma^2$ and $L$ into $\lambda$.

We also studied the effect of learning $\lambda$ when the weights of encoder/decoder are not tied, similar to~\cite{SreterHillel2018LCSC}. We call the resulting architectures CRsAE-$\lambda_{\text{EM}}$-untied and CRsAE-$\lambda_{\text{LS}}$-untied. For both cases, the encoder had $T = 30$ layers. We studied the effect of the depth of encoder unfolding by training an architecture, CRsAE-$\lambda_{\text{EM}}$-shallow, for which $T=3$. At last, to emphasize the significance of the method we propose to train $\lambda$, we trained CRsAE-$\lambda_{\text{EM}}$-untied to minimize a combination of the $\ell_1$ and multiscale structural similarity~\cite{wang2003multiscale} (MS-SSIM) losses, similar to~\cite{SreterHillel2018LCSC}. We call this network CRsAE-$\lambda_{\text{EM}}$-untied-msssim. In this case, the main difference between LCSC and CRsAE is the training procedure.

We initialized the filters using i.i.d. draws from a standard Gaussian random variable. We trained the network for $200$ epochs. At every iteration, we cropped a $128 \times 128$ patch from the training image and added random Gaussian noise with standard deviation $\frac{20}{255}$. We set the initial learning rate $\eta_{\smlh}$ to $10^{-2}$ and decreased it by a factor of $0.7$ every $10$ epochs. We set $\eta_{\lambda}$ to $0.1$ for EM and $0.01$ for the LS method. We summarize the parameters used for training in Table~\ref{tab:dataset2d}.

\noindent Table~\ref{tab:psnr} presents the performances on the denoising task.

\noindent \underline{\textbf{Encoder depth improves denoising}}: Comparing CRsAE-$\lambda_{\text{EM}}$-shallow and CRsAE-$\lambda_{\text{EM}}$ suggests that having a deep encoder can significantly improve performance--in this case, by $1$-$1.5$ dB in PSNR.

\noindent \underline{\textbf{EM-inspired training improves denoising}}: We can draw three conclusions by comparing CRsAE-$\lambda_{\text{hyp}}$, CRsAE-$\lambda_{\text{LS}}$, and CRsAE-$\lambda_{\text{EM}}$. First, training $\lambda$ by grid search results in poorer performance compared to training it using the standard approach or the EM-inspired approach. Second, and more importantly, the EM-inspired method outperforms the standard approach. This behaviour persists for the untied case as well. Indeed, the performance of CRsAE-$\lambda_{\text{EM}}$-untied-msssim rivals that of LCSC~\cite{SreterHillel2018LCSC}, which trains $\lambda$ by minimizing a global reconstruction loss.

\noindent \underline{\textbf{Depth and weight-tying affect learned dictionary}}: Figure~\ref{fig:lena} depcits the result of denoising the test image (Lena) using CRsAE-$\lambda_{\text{EM}}$-shallow and CRsAE-$\lambda_{\text{EM}}$-untied-msssim. Figure~\ref{fig:filters} shows the filters that each of the networks learned. The figure demonstrates that a network with a deep encoder, and for which the encoder and decoder share weigths, is better able to learn filters that are Gabor-like edge detectors~\cite{MehrotraR1992Gfed}, compared to a network with a shallow encoder or a network whose encoder and decoder do not share weights. 

%%%%%%%%%%%%%%%%%%%%%%%%%%%%%%%%%%%%%%%%%%%%%%%%%%%%%%%%%%%%%%%%%%%%%%
\vspace*{-3mm}
\section{Conclusion}
\label{sec:CONC}
Training an autoencoder for convolutional sparse coding to result in very low input/output prediction error is not a challenging task and has been done before~\cite{SreterHillel2018LCSC}. The challenging problem is to demonstrate that the weights learned by an autoencoder are interpretable as convolutional filters in the context of convolutional dictionary learning or source separation, a problem that, to our knowledge, existing autoencoder architectures for convolutional sparse coding have not solved. In this paper, we framed convolutional dictionary learning as training of an autoencoder and proposed the constrained recurrent sparse autoencoder (CRsAE). CRsAE is the first autoencoder architecture to learn interpretable filters corresponding to a generative model. We highlighted this ability of CRsAE and its robustness to noise through simulation. CRsAE owes it success at performing dictionary learning and source separation to a) the deep recurrent and residual neural network architecture, inherited from FISTA, of its encoder, b) the tying of the weights to be learned in the encoder and decoder, and c) the Expectation Maximization-inspired training of the regularization parameter to ensure the sparseness of codes output by the encoder. The tying constraint on the encoder and decoder reduces the trainable parameters by a factor of $\frac{1}{3}$. In addition, the recurrent architecture of CRsAE with shared weights allows us to produce sparse representations as the depth increases with a fixed number of trainable parameters. 

We highlighted the similarities and differences of methods such as classical dictionary learning, Expectation Maximization, and CRsAE for learning sparse representations from data. The architecture for CRsAE and the training procedure are motivated by the Expectation Maximization algorithm. The encoder in CRsAE produces sparse codes, a similar step to the sparse coding update in dictionary learning, and the E-step of Expectation Maximization. CRsAE learns the filters and the regularization parameter using a two-stage back-propagation procedure, a step corresponding to the M-step in Expectation Maximization. In the first stage, the filters of interest are trained by back-propagation through the autoencoder. This step parallels the dictionary update step in convolutional dictionary learning. In the second stage, the regularization parameter is updated by back-propagation through the encoder using a loss function motivated by Bayesian statistics. This two-stage back-propagation and the loss function for training the regularization parameter are the keys to ensuring the sparseness of the codes, output by the encoder, and preventing the convergence of the regularization parameter (bias) to zero.

CRsAE applies to the case when the posterior concentrates around the MAP estimate, an assumption that is inherited from its parallel with the alternating-minimization algorithm for dictionary learning. An interesting line of work is to extend CRsAE to cases when this assumption fails. In such cases, the output of the encoder ought to approximate samples from the posterior distribution of the codes. The resulting architecture would resemble variational autoencoders~\cite{KingmaW13,chen2016variational}, in the sense that the latent representation is random. Moreover, the encoder and decoder would be tied as in CRsAE.

We benchmarked CRsAE and showed that it performs dictionary learning $5$x faster than the state-of-the-art algorithm~\cite{garcia-2018-convolutional} for convolutional dictionary learning based on alternating-minimization. We showed that the encoder of CRsAE could identify the location of action potentials in extracellular voltage recordings from the brain of rats. In particular,  we showed that CRsAE could perform spike sorting as well as the state-of-the-art algorithm based on convex optimization known as continuous basis pursuit. At the same time, compared to continuous basis pursuit, CRsAE significantly speeds up the process of spike sorting by $900$x.

Finally, we demonstrated the performance of CRsAE on the task of image denoising. We showed that, when learning the regularization parameter through the proposed EM-inspired approach, CRsAE denoises images better than networks that train the regularization parameter (ReLU bias) by minimizing a global reconstruction loss, the standard approach in deep learning. We also showed that the performance of our method rivals that of LCSC~\cite{SreterHillel2018LCSC}. At last, we highlighted that both a deep encoder and the sharing of weights between the encoder and decoder promote learning of Gabor-like filters~\cite{MehrotraR1992Gfed}.

% acknowledgment
\vspace*{-2.5mm}
\section*{Acknowledgments}

\noindent The authors gratefully acknowledge supports by NSF Simons Center for Mathematical and Statistical Analysis of Biology at Harvard University (grant no. DMS-1764269), the Harvard FAS Quantitative Biology Initiative. This research is also supported by AWS Machine Learning Research Awards.

\appendix  % for no appendix heading

\noindent Here, we derive the back-propagation algorithm for CRsAE. This derivation is only for completeness. In practice, the gradients are computed through autograd function while training. We assume, without loss of generality, number of examples $J=1$. Let $\smlh = [\smlh_1^\text{T},\ldots,\smlh_C^\text{T}]^\text{T}$ where $\smlh_c = [h_c[0],h_c[1],\ldots,h_c[K-1]]^{\text{T}}$, and $\w_t = [\w_{t,1}^\text{T},\ldots,\w_{t,C}^\text{T}]^\text{T}$ where $\w_{t,c} = [w_{t,c}[0],w_{t,c}[1],\ldots,w_{t,c}[N_e-1]]^{\text{T}}$. Let $\cvec_{T+1} = \hat \y = \Bigh \cvec_T$. The gradient of the loss function in Eq. (27) from the main paper is
\begin{equation}
	\frac{\partial \mathcal{L}_H}{\partial \smlh} = \delta \smlh = \sum_{t=1}^{T+1}\frac{\partial \cvec_t}{\partial \smlh} \frac{\partial \mathcal{L}_H}{\partial \cvec_t} = \sum_{t=1}^{T+1} \frac{\partial \cvec_t}{\partial \smlh} \delta \cvec_t,
	\label{eq:partialh}
\end{equation}
\noindent where
\begin{equation}
	\delta \cvec_t = \frac{\partial \z_t}{\partial \cvec_t} \frac{\partial \mathcal{L}_H}{\partial \z_t} = \frac{\partial \z_t}{\partial \cvec_t}  \delta \z_t.
	\label{eq:particalct}
\end{equation}
\noindent To evaluate Eq.~(\ref{eq:particalct}), we first compute $\frac{\partial \z_{t}}{\partial \cvec_{t}}$ as
\begin{equation}
  \frac{\partial \z_{t}}{\partial \cvec_{t}}=
               \begin{bmatrix}
               	\text{diag}(\prox_{b}'(\cvec_t))| \mathbf{0}_{N_eC} \in \R^{N_eC \times 2N_eC}
               \end{bmatrix},
  \label{eq:partialz_c}
\end{equation}

\noindent where $b = \frac{\lambda \sigma^2}{L}$. Then, we evaluate $\delta \z_{t-1}$ through the following recursion
\begin{equation}
	\delta \z_{t-1} = \frac{\partial \z_t}{\partial \z_{t-1}} \frac{\partial \mathcal{L}_H}{\partial \z_t} = \frac{\partial \cvec_t}{\partial \z_{t-1}}\frac{\partial \z_t}{\partial \cvec_{t}} \frac{\partial \mathcal{L}_H}{\partial \z_t} = \frac{\partial \cvec_t}{\partial \z_{t-1}} \delta \cvec_t,
	\label{eq:partialzt1}
\end{equation}

Having $\cvec_{T+1} = \hat \y$, we initialize the recursion with
\begin{equation}
	\delta \cvec_{T+1} = \frac{\partial \mathcal{L}_H}{\partial \cvec_{T+1}} = \frac{\partial \hat \y}{\partial \cvec_{T+1}} \frac{\partial \mathcal{L}_H}{\partial \hat \y} = \eye_{N} \delta \hat \y = \hat \y-\y.
	\end{equation}
	
\noindent We then evaluate $\frac{\partial \cvec_{t}}{\partial \z_{t-1}}$ needed for Eq.~(\ref{eq:partialzt1}) as
\[
  \frac{\partial \cvec_{t}}{\partial \z_{t-1}}=\begin{cases}
                              \begin{bmatrix}
               	\Bigh^{\text{T}} \\ \mathbf{0}_{N_eC\times N}
               \end{bmatrix} \in \R^{2N_eC \times N}, \text{if } t=T+1\\
               \begin{bmatrix}
               	\left(1 + \frac{s_{t-1}-1}{s_t}\right) \eye_{N_eC} \\  -\frac{s_{t-1}-1}{s_t} \eye_{N_eC}
               \end{bmatrix}
               \begin{bmatrix}
               	\eye_{N_eC} - \frac{1}{L} \Bigh^{\text{T}}\Bigh.
               \end{bmatrix}
               \text{,otherwise}.
            \end{cases}
\]
	
\noindent Finally, to finish the full gradient propagation procedure, we compute $\frac{\partial \cvec_{t}}{\partial \smlh}$ needed for Eq.~(\ref{eq:partialh}) as 

\begin{equation}
	\frac{\partial \cvec_{T+1}}{\partial \smlh} = \mathbf{Z}_{T\ast}^{(1)} \in \R^{KC \times N}.
\end{equation}
\noindent where
\begin{equation}
\left[
    \begin{array}{c}
    \mathbf{Z}_{T\ast,1}^{(1)}\\ \hdashline[1pt/5pt]
    \mathbf{Z}_{T\ast,2}^{(1)} \\ \hdashline[1pt/5pt]
    \vdots \\ \hdashline[1pt/5pt]
    \mathbf{Z}_{T\ast,C}^{(1)} \\
\end{array} \right]
\end{equation}
\vspace*{-1.0mm}
\begin{equation}
\mathbf{Z}_{T\ast,c}^{(1)} = \begin{bmatrix}
    x_{T,c}^{(1)}[0] & x_{T,c}^{{(1)}}[1] & x_{T,c}^{(1)}[2] & \dots  & \dots \\
    0 & x_{T,c}^{(1)}[0] & x_{T,c}^{(1)}[1] & \dots  & \dots \\
    \vdots & \vdots & x_{T,c}^{(1)}[0] & \ddots & \vdots \\
    0 & 0 & 0 & \dots  & \dots
\end{bmatrix}_{K\times N}
\end{equation}

\noindent and for $t = 1, \ldots, T$,
\begin{equation}
\begin{aligned}
&\left(\frac{\partial \cvec_t}{\partial \smlh}\right)_{(a-1)K+m,(b-1)N_e+n} =\\
\frac{1}{L} \sum_{\{|k|< K\}} \sum_{g=1}^C & w_{t,g}[n-1+k] \frac{\partial}{\partial h_{a}[m-1]} c_{\smlh_b\smlh_g}[k]\\
+& y[m+n-2] \mathbb{I}_{\{a=b\}}
\end{aligned}
\end{equation}

\noindent for $a,b = 1,\ldots, C$, $m=1,\ldots,K$, and $n=1,\ldots,N_e$. $c_{\smlh_b\smlh_g}[k]$ is the deterministic cross-correlation function between $\smlh_b$ and $k$-delayed samples of $\smlh_g$ as follows
\begin{equation}
\vspace*{-2.0mm}
\begin{aligned}
c_{\smlh_b\smlh_g}[k] = \sum_{n=0}^{K-k-1} \smlh_b[n+k]  \smlh_g[n]
\end{aligned}
\end{equation}

Next, we derive the back-propagation algorithm for CRsAE for training $\lambda$. Without loss of generality, we assume $\sigma = 1$. The gradient of the loss function in Eq. (28) from the main
paper w.r.t. $\lambda$ is
\begin{equation}
\begin{aligned}
	\frac{\partial \mathcal{L}_{\lambda}}{\partial \lambda} &= (\|\z_T^{(1)}\|_1 + C\delta + \lambda \frac{\partial \|\z_T^{(1)}\|_1}{\partial \lambda}) - \frac{(N_e + r - 1)C}{\lambda}\\
	&= \|\z_T^{(1)}\|_1 + C\delta - \frac{(N_e+ r - 1)C}{\lambda} + \lambda \sum_{t=1}^{T}\frac{\partial \z_t}{\partial \lambda} \frac{\partial \|\z_T^{(1)}\|_1}{\partial \z_t}
	\label{eq:partiallambda}
\end{aligned}
\end{equation}
\vspace*{-2.0mm}
\noindent We evaluate  $\frac{\partial \|\z_T^{(1)}\|_1}{\partial \z_{t-1}}$ through the following recursion
\begin{equation}
\vspace*{-2.0mm}
	\frac{\partial \|\z_T^{(1)}\|_1}{\partial \z_{t-1}} = \frac{\partial \z_t}{\partial \z_{t-1}} \frac{\partial  \|\z_T^{(1)}\|_1}{\partial \z_t}
	\label{eq:partialzt1zt}
\end{equation}
We initialize the recursion with
\begin{equation}
\left(\frac{\partial  \|\z_T^{(1)}\|_1}{\partial \z_{T}}\right)_i = \begin{cases}
	\text{sign}(\z_{T}[i])\ \text{if}\ \z_{T}[i] \neq 0\\
	0\quad \text{,otherwise}
\end{cases}
\end{equation}
\noindent where $\frac{\partial  \|\z_T^{(1)}\|_1}{\partial \z_{T}} \in \R^{2N_eC \times 1}$ (Note that for when $\z_{T}[i] = 0$, we have used the sub-gradient as the gradient is not defined). Then we evaluate
\begin{equation}
\vspace*{-2.0mm}
	 \frac{\partial \z_t}{\partial \z_{t-1}} =  \frac{\partial \z_t}{\partial \cvec_{t}}  \frac{\partial \cvec_t}{\partial \z_{t-1}}
\end{equation}
\noindent where $\frac{\partial \z_{t}}{\partial \cvec_{t}}$ and $\frac{\partial \cvec_{t}}{\partial \z_{t-1}}$ are derived previously in the derivation of back-propagation for the filters.
	
Finally, to finish the full gradient propagation procedure, we compute $\frac{\partial \z_{t}}{\partial \lambda} \in \R^{1 \times 2N_eC}$ needed for Eq.~(\ref{eq:partiallambda}) from $\z_t^{(1)} = \prox_{b}(\cvec_t)$
\vspace*{-2.0mm}
\begin{equation}
	\left(\frac{\partial \z_{t}}{\partial \lambda}\right)_i = \begin{cases} 
 	-\frac{\sigma^2}{L}\ \text{sign}(c_{t}[i])\ \text{if}\ |c_{t}[i]| \geq b\\
 	0 \text{, otherwise.}	
 	 \end{cases}
	 \vspace*{-2.0mm}
\end{equation}

\vspace*{-2.0mm}
\ifCLASSOPTIONcaptionsoff
  \newpage
\fi

\bibliographystyle{IEEEtran.bst}
\bibliography{IEEEfull,TNNLS-2019-P-11258.R2.bib}

\begin{thebibliography}{10}
\providecommand{\url}[1]{#1}
\csname url@samestyle\endcsname
\providecommand{\newblock}{\relax}
\providecommand{\bibinfo}[2]{#2}
\providecommand{\BIBentrySTDinterwordspacing}{\spaceskip=0pt\relax}
\providecommand{\BIBentryALTinterwordstretchfactor}{4}
\providecommand{\BIBentryALTinterwordspacing}{\spaceskip=\fontdimen2\font plus
\BIBentryALTinterwordstretchfactor\fontdimen3\font minus
  \fontdimen4\font\relax}
\providecommand{\BIBforeignlanguage}[2]{{%
\expandafter\ifx\csname l@#1\endcsname\relax
\typeout{** WARNING: IEEEtran.bst: No hyphenation pattern has been}%
\typeout{** loaded for the language `#1'. Using the pattern for}%
\typeout{** the default language instead.}%
\else
\language=\csname l@#1\endcsname
\fi
#2}}
\providecommand{\BIBdecl}{\relax}
\BIBdecl

\bibitem{Chen1989OrthogonalLS}
S.~Chen, S.~A. Billings, and W.~Luo, ``Orthogonal least squares methods and
  their application to non-linear system identification,'' \emph{International
  Journal of Control}, vol.~50, no.~5, pp. 1873--96, 1989.

\bibitem{Chen1998AtomicDB}
S.~S. Chen, D.~L. Donoho, and M.~A. Saunders, ``Atomic decomposition by basis
  pursuit,'' \emph{SIAM Review}, vol.~43, pp. 129--59, 1998.

\bibitem{TibshiraniRobert1996RSaS}
R.~Tibshirani, ``Regression shrinkage and selection via the lasso,''
  \emph{Journal of the Royal Statistical Society: Series B (Methodological)},
  vol.~58, no.~1, pp. 267--88, 1996.

\bibitem{DempsterAP1977MLfI}
A.~P. Dempster, N.~M. Laird, and D.~B. Rubin, ``Maximum likelihood from
  incomplete data via the em algorithm,'' \emph{Journal of the Royal
  Statistical Society: Series B (Methodological)}, vol.~39, no.~1, pp. 1--22,
  1977.

\bibitem{ZhuXinFeng2011LSLa}
X.~F. Zhu, B.~Li, and J.~D. Wang, ``\BIBforeignlanguage{eng}{L1-norm sparse
  learning and its application},'' \emph{\BIBforeignlanguage{eng}{Applied
  Mechanics and Materials}}, vol. 88-89, pp. 379--85, 2011.

\bibitem{ParkTrevor2008TBL}
T.~Park and G.~Casella, ``\BIBforeignlanguage{eng}{The bayesian lasso},''
  \emph{\BIBforeignlanguage{eng}{Journal of the American Statistical
  Association}}, vol. 103, no. 482, pp. 681--86, 2008.

\bibitem{Mallat1989ATF}
S.~Mallat, ``A theory for multiresolution signal decomposition: The wavelet
  representation,'' \emph{IEEE Transactions on Pattern Analysis and Machine
  Intelligence}, vol.~11, pp. 674--693, 1989.

\bibitem{EnganK1999Mood}
K.~Engan, S.~Aase, and J.~Hakon~Husoy, ``\BIBforeignlanguage{eng}{Method of
  optimal directions for frame design},'' in
  \emph{\BIBforeignlanguage{eng}{Proc. IEEE International Conference on
  Acoustics, Speech, and Signal Processing. Proceedings (ICASSP)}}, vol.~5,
  1999, pp. 2443--46.

\bibitem{AharonM2006rKAA}
M.~Aharon, M.~Elad, and A.~Bruckstein, ``\BIBforeignlanguage{eng}{K-svd: An
  algorithm for designing overcomplete dictionaries for sparse
  representation},'' \emph{\BIBforeignlanguage{eng}{IEEE Transactions on Signal
  Processing}}, vol.~54, no.~11, pp. 4311--22, 2006.

\bibitem{garcia-2018-convolutional}
C.~Garcia-Cardona and B.~Wohlberg, ``Convolutional dictionary learning: A
  comparative review and new algorithms,'' \emph{IEEE Transactions on
  Computational Imaging}, vol.~4, no.~3, pp. 366--81, Sep. 2018, there are
  errors in Equations (18) and (19) in the published version of the paper.
  These have been corrected in the most recent arXiv version.

\bibitem{PapyanV2017CNNA}
V.~Papyan, Y.~Romano, and M.~Elad, ``\BIBforeignlanguage{English}{Convolutional
  neural networks analyzed via convolutional sparse coding},''
  \emph{\BIBforeignlanguage{English}{Journal of Machine Learning Research}},
  vol.~18, pp. 1--52, 2017.

\bibitem{Agarwal2016LearningSU}
A.~Agarwal, A.~Anandkumar, P.~Jain, P.~Netrapalli, and R.~Tandon, ``Learning
  sparsely used overcomplete dictionaries via alternating minimization,''
  \emph{SIAM Journal on Optimization}, vol.~26, pp. 2775--99, 2016.

\bibitem{Boyd2011DistributedOA}
S.~P. Boyd, N.~Parikh, E.~Chu, B.~Peleato, and J.~Eckstein, ``Distributed
  optimization and statistical learning via the alternating direction method of
  multipliers,'' \emph{Foundations and Trends in Machine Learning}, vol.~3, pp.
  1--122, 2011.

\bibitem{SreterHillel2018LCSC}
H.~Sreter and R.~Giryes, ``\BIBforeignlanguage{eng}{Learned convolutional
  sparse coding},'' in \emph{\BIBforeignlanguage{eng}{Proc. IEEE International
  Conference on Acoustics, Speech and Signal Processing (ICASSP)}}, 2018, pp.
  2191--95.

\bibitem{Rolfe2013DiscriminativeRS}
J.~T. Rolfe and Y.~LeCun, ``Discriminative recurrent sparse auto-encoders,'' in
  \emph{Proc. International Conference on Learning Representations (ICLR)},
  2013, pp. 1--15.

\bibitem{Makhzani2013kSparseA}
A.~Makhzani and B.~J. Frey, ``k-sparse autoencoders,'' in \emph{Proc.
  International Conference on Learning Representations (ICLR)}, 2014, pp. 1--9.

\bibitem{TolooshamsBahareh2018SCDL}
B.~Tolooshams, S.~Dey, and D.~Ba, ``\BIBforeignlanguage{eng}{Scalable
  convolutional dictionary learning with constrained recurrent sparse
  auto-encoders},'' in \emph{\BIBforeignlanguage{eng}{Proc. IEEE 28th
  International Workshop on Machine Learning for Signal Processing (MLSP)}},
  2018, pp. 1--6.

\bibitem{Blumensath2008IterativeHT}
T.~Blumensath and M.~E. Davies, ``Iterative hard thresholding for compressed
  sensing,'' \emph{Applied and Computational Harmonic Analysis}, vol.~27,
  no.~3, pp. 265--74, 2009.

\bibitem{HeKaiming2015DRLf}
K.~{He}, X.~{Zhang}, S.~{Ren}, and J.~{Sun}, ``Deep residual learning for image
  recognition,'' in \emph{Proc. IEEE Conference on Computer Vision and Pattern
  Recognition (CVPR)}, 2016, pp. 770--8.

\bibitem{gregor2010learning}
K.~Gregor and Y.~LeCun, ``Learning fast approximations of sparse coding,'' in
  \emph{Proc. the 27th International Conference on Machine Learning (ICML)},
  2010, pp. 399--406.

\bibitem{Zeiler2010DeconvolutionalN}
M.~D. {Zeiler}, D.~{Krishnan}, G.~W. {Taylor}, and R.~{Fergus},
  ``Deconvolutional networks,'' in \emph{Proc. IEEE Computer Society Conference
  on Computer Vision and Pattern Recognition (CVPR)}, 2010, pp. 2528--35.

\bibitem{Donoho2197}
D.~L. Donoho and M.~Elad, ``Optimally sparse representation in general
  (nonorthogonal) dictionaries via l1 minimization,'' \emph{Proc. the National
  Academy of Sciences}, vol. 100, no.~5, pp. 2197--2202, 2003.

\bibitem{lewicki1998review}
M.~S. Lewicki, ``A review of methods for spike sorting: the detection and
  classification of neural action potentials,'' \emph{Network: Computation in
  Neural Systems}, vol.~9, no.~4, pp. R53--R78, 1998.

\bibitem{Cands2005StableSR}
E.~J.~Cand{\`e}s, J.~K.~Romberg, and T.~Tao, ``Stable signal recovery from
  incomplete and inaccurate measurements,'' \emph{Communications on Pure and
  Applied Mathematics}, vol.~59, pp. 1--15, 2006.

\bibitem{Cands2008TheRI}
E.~J. Cand{\`e}s, ``The restricted isometry property and its implications for
  compressed sensing,'' \emph{Comptes Rendus Mathematique}, vol. 346, no.~9,
  pp. 589--92, 2008.

\bibitem{wohlberg-2014-efficient}
B.~Wohlberg, ``Efficient convolutional sparse coding,'' in \emph{Proc. IEEE
  International Conference on Acoustics, Speech, and Signal Processing
  (ICASSP)}, Florence, Italy, 2014, pp. 7173--77.

\bibitem{Daubechies2003AnIT}
I.~Daubechies, M.~Defrise, and C.~De~Mol, ``An iterative thresholding algorithm
  for linear inverse problems with a sparsity constraint,''
  \emph{Communications on Pure and Applied Mathematics: A Journal Issued by the
  Courant Institute of Mathematical Sciences}, vol.~57, no.~11, pp. 1413--57,
  2004.

\bibitem{beck2009fast}
A.~Beck and M.~Teboulle, ``A fast iterative shrinkage-thresholding algorithm
  for linear inverse problems,'' \emph{SIAM journal on imaging sciences},
  vol.~2, no.~1, pp. 183--202, 2009.

\bibitem{HersheyJohnR2014DUMI}
J.~R. Hershey, J.~L. Roux, and F.~Weninger, ``Deep unfolding: Model-based
  inspiration of novel deep architectures,'' \emph{arXiv:1409.2574 [cs.LG]},
  pp. 1--27, 2014.

\bibitem{mardani2018}
M.~Mardani, Q.~Sun, D.~Donoho, V.~Papyan, H.~Monajemi, S.~Vasanawala, and
  J.~Pauly, ``Neural proximal gradient descent for compressive imaging,'' in
  \emph{Proc. Advances in Neural Information Processing Systems 31}, S.~Bengio,
  H.~Wallach, H.~Larochelle, K.~Grauman, N.~Cesa-Bianchi, and R.~Garnett,
  Eds.\hskip 1em plus 0.5em minus 0.4em\relax Curran Associates, Inc., 2018,
  pp. 9573--83.

\bibitem{rethinking}
D.~Simon and M.~Elad, ``Rethinking the csc model for natural images,'' in
  \emph{Proc. Advances in Neural Information Processing Systems 33 (NeurIPS)},
  2019.

\bibitem{KingmaW13}
\BIBentryALTinterwordspacing
D.~P. Kingma and M.~Welling, ``Auto-encoding variational bayes,'' in \emph{2nd
  International Conference on Learning Representations, {ICLR}}, 2014.
  [Online]. Available: \url{http://arxiv.org/abs/1312.6114}
\BIBentrySTDinterwordspacing

\bibitem{chen2016variational}
X.~Chen, D.~P. Kingma, T.~Salimans, Y.~Duan, P.~Dhariwal, J.~Schulman,
  I.~Sutskever, and P.~Abbeel, ``Variational lossy autoencoder,'' \emph{arXiv
  preprint arXiv:1611.02731}, 2016.

\bibitem{Kingma2014AdamAM}
D.~P. Kingma and J.~Ba, ``Adam: A method for stochastic optimization,'' in
  \emph{Proc. the 3rd International Conference on Learning Representations
  (ICLR)}, 2014, pp. 1--15.

\bibitem{Smith2017CyclicalLR}
L.~N. Smith, ``Cyclical learning rates for training neural networks,'' in
  \emph{Proc. IEEE Winter Conference on Applications of Computer Vision
  (WACV)}, vol.~1, 2017, pp. 464--72.

\bibitem{KrizhevskyAlex2017Icwd}
A.~Krizhevsky, I.~Sutskever, and G.~Hinton, ``Imagenet classification with deep
  convolutional neural networks,'' in \emph{Proc. Advances in Neural
  Information Processing Systems 25 (NIPS)}, 2017, pp. 1097--105.

\bibitem{GlorotXavier2011DSRN}
X.~Glorot, A.~Bordes, and Y.~Bengio, ``Deep sparse rectifier neural networks,''
  in \emph{Proc. the 14th International Conference on Artificial Intelligence
  and Statistics}, 2011, pp. 315--23.

\bibitem{wohlberg-2017-sporco}
B.~Wohlberg, ``{SPORCO}: {A} {P}ython package for standard and convolutional
  sparse representations,'' in \emph{Proc. 15th Python in Science Conference},
  Austin, TX, USA, 2017, pp. 1--8.

\bibitem{HarrisKd2000Aots}
K.~Harris, D.~Henze, J.~Csicsvari, H.~Hirase, and G.~Buzsaki,
  ``\BIBforeignlanguage{English}{Accuracy of tetrode spike separation as
  determined by simultaneous intracellular and extracellular measurements},''
  \emph{\BIBforeignlanguage{English}{Journal Of Neurophysiology}}, vol.~84,
  no.~1, pp. 401--14, 2000.

\bibitem{ekanadham2014unified}
C.~Ekanadham, D.~Tranchina, and E.~P. Simoncelli, ``A unified framework and
  method for automatic neural spike identification,'' \emph{Journal of
  neuroscience methods}, vol. 222, pp. 47--55, 2014.

\bibitem{LecunY1998Eb}
Y.~A. LeCun, L.~Bottou, G.~B. Orr, and K.-R. M{\"u}ller,
  ``\BIBforeignlanguage{English}{Efficient backprop},''
  \emph{\BIBforeignlanguage{English}{Neural Networks: Tricks Of The Trade}},
  vol. 1524, pp. 9--50, 1998.

\bibitem{pascal-voc-2012}
M.~Everingham, L.~Van~Gool, C.~K.~I. Williams, J.~Winn, and A.~Zisserman, ``The
  {PASCAL} {V}isual {O}bject {C}lasses {C}hallenge 2012 {(VOC2012)}
  {R}esults,'' 2012.

\bibitem{wang2003multiscale}
Z.~Wang, E.~P. Simoncelli, and A.~C. Bovik, ``Multiscale structural similarity
  for image quality assessment,'' in \emph{The Thrity-Seventh Asilomar
  Conference on Signals, Systems \& Computers, 2003}, vol.~2.\hskip 1em plus
  0.5em minus 0.4em\relax Ieee, 2003, pp. 1398--1402.

\bibitem{MehrotraR1992Gfed}
R.~Mehrotra, K.~Namuduri, and N.~Ranganathan, ``\BIBforeignlanguage{eng}{Gabor
  filter-based edge detection},'' \emph{\BIBforeignlanguage{eng}{Pattern
  Recognition}}, vol.~25, no.~12, pp. 1479--94, 1992.

\end{thebibliography}

\end{document}